\documentclass[twocolumn,10pt,a4paper]{article}

\usepackage[T1]{fontenc}
\usepackage[utf8]{inputenc}
\usepackage[english]{babel}
\usepackage[default]{sourceserifpro}
\usepackage[scaled=0.93]{sourcesanspro}
\usepackage[scaled=0.85]{sourcecodepro}

\usepackage[margin=2cm,top=2cm,bottom=2cm,columnsep=18pt]{geometry}

\usepackage{microtype}
\usepackage{xcolor}
\definecolor{linkblue}{HTML}{1F4E79}

\usepackage{titlesec}
\titleformat{\section}
  {\normalfont\large\bfseries\sffamily}{\thesection.}{0.5em}{\MakeUppercase}
  [\vspace{1pt}\titlerule]
\titleformat{\subsection}
  {\normalfont\normalsize\bfseries\sffamily}{\thesubsection.}{0.5em}{}
\titlespacing*{\section}{0pt}{0.6em}{0.3em}
\titlespacing*{\subsection}{0pt}{0.4em}{0.2em}
\titlespacing*{\paragraph}{0pt}{0.3em}{0.5em}

\usepackage{graphicx}
\usepackage{booktabs}
\usepackage{multirow}
\usepackage{array}
\usepackage{tabularx}
\usepackage{float}
\usepackage[font=small,labelfont=bf,labelsep=period,skip=4pt]{caption}

\usepackage{amsmath,amssymb}
\usepackage{url}
\usepackage{hyperref}
\usepackage{enumitem}
\usepackage{tikz}
\usetikzlibrary{positioning,arrows.meta,calc,fit,backgrounds}
\usepackage{placeins}
\usepackage{afterpage}
\usepackage{balance}
\hypersetup{colorlinks=true,linkcolor=linkblue,urlcolor=linkblue,citecolor=linkblue}

\graphicspath{{figures/}}
\newcommand{\mdlogo}[1]{\raisebox{-.3ex}{\includegraphics[height=1.05em]{logos/#1_logo}}}

\title{}\author{}\date{}

\begin{document}
\thispagestyle{empty}

\twocolumn[%
  \begin{@twocolumnfalse}%
    \vspace{12pt}
    \begin{center}
      {\fontfamily{ptm}\selectfont\Huge\bfseries Brick: Spatial Capability Routing for the Mixture-of-Models (MoM) Paradigm}\par
      \vspace{8pt}
      {\fontfamily{ptm}\selectfont\Large\color{black!75} Bridging Heterogeneous LLM Pools at Inference Time}\par
      \vspace{14pt}
      {\Large\bfseries May 2026}\par
      \vspace{12pt}
      \begin{tabular}{c@{\hspace{3em}}c}
        {\large Francesco Massa} & {\large Marco Cristofanilli} \tabularnewline
        {\small \href{mailto:f.massa@regolo.ai}{f.massa@regolo.ai}} &
        {\small \href{mailto:marco.c@seeweb.it}{marco.c@seeweb.it}} \tabularnewline
      \end{tabular}\par
      \vspace{8pt}
      {\normalsize \href{https://github.com/regolo-ai/brick-SR1}{regolo-ai/brick-SR1} \quad\textbullet\quad \href{https://huggingface.co/regolo}{huggingface.co/regolo}}
      \par\vspace{6pt}
      {\small\color{black!60} Technical Report \textbullet{} v1.0 \textbullet{} \href{https://creativecommons.org/licenses/by/4.0/}{CC BY 4.0}}
    \end{center}
    \vspace{12pt}

    \section*{Abstract}
    \emph{What is difficult for a language model?} Defining difficulty is one of the hardest problems in deployment engineering, and the only reliable answer is empirical. Once correctness is measured per query and per model, the question becomes: can a router dispatch each query to the cheapest model that will answer it correctly?

    \vspace{4pt}\noindent \textbf{The problem with existing answers.} Existing systems route on surface features: domain labels, keywords, token count. We call this \emph{superficial routing}. It collapses the within-domain variance that actually determines which model succeeds, since a one-line question can target an unsolved problem while a long boilerplate prompt can wrap a trivial task.

    \vspace{4pt}\noindent \textbf{Why this matters at scale.} Frontier commercial models charge ten to one hundred times what local open-weight models cost. At production scale even small per-request savings compound across millions of calls into a direct cloud-bill lever, and the bill is the number operators are measured on.

    \vspace{4pt}\noindent \textbf{What we do.} We present \textbf{Brick}, a multimodal router that scores each model on six capability dimensions, combines this with a per-query difficulty estimate, and dispatches via a cost-penalized geometric rule. A continuous preference knob $r$ lets operators slide between a \emph{max-quality} and a \emph{max-saving} profile, turning the trade-off into a deploy-time parameter rather than a code change.

    \vspace{4pt}\noindent \textbf{What follows.} On \textbf{Dataset A} ($5{,}504$ queries, $\kappa{=}0.761$), Brick at max-quality reaches $\mathbf{76.98\%}$ accuracy, beating the best single model (\texttt{kimi2.6}, $75.02\%$) and all external routers (RouteLLM, FrugalGPT, Cascade Routing). The cost-quality knob exposes a continuous trade-off: at the low profile Brick beats FrugalGPT's $69.42\%$ by $+2.20$\,pp at comparable cost, at the neutral profile Brick reaches $\mathbf{74.11\%}$ at $\mathbf{4.71{\times}}$ lower cost than always-\texttt{kimi} (only $0.91$\,pp accuracy loss), and at the min-cost profile it cuts cost $\mathbf{22.15{\times}}$ at $11.85$\,pp accuracy loss. The neutral profile also beats Cascade Routing's $73.40\%$ by $+0.71$\,pp. Median end-to-end latency drops from $51.2$\,s to $22.8$\,s; the three-model oracle bound is $83.25\%$.

    \medskip
    \paragraph{Keywords.} LLM routing, mixture of models, capability classification, cost-aware inference, open-weight models.
    \vspace{10pt}
  \end{@twocolumnfalse}%
]

\section{Introduction}\label{sec:intro}
Identifying what is \emph{difficult} for a given language model is one of the harder skills a deployment engineer can develop. Training data, architecture changes, and reinforcement post-training all interact in opaque ways, so deciding which model is best for a task is in practice an empirical exercise: only repeated evaluation reveals where a model excels and where it fails silently.

Consider two queries that both arrive at the API with the same \texttt{Content-Type} header. The first asks for a Python function that returns the $n$-th Fibonacci number; the second asks for a proof that the Riemann zeta function has no non-trivial zeros outside the critical strip. Every model in our pool answers the first correctly, including a 9B-parameter open-weight one at $\$0.04$/$\$0.15$ per $10^6$ input/output tokens; none of them solves the second. Dispatching both to a frontier model wastes more than $20{\times}$ the budget on the first and the same budget plus minutes of latency on the second. The router's job is to know, per query, which model is the cheapest among those that will succeed.

A growing collection of public evaluation suites makes this exercise tractable. Once response correctness can be measured per query and per model, the question becomes: given a pool of $K$ candidate models with heterogeneous price and capability profiles, can a router send each query to the cheapest model that will answer it correctly? This is the routing problem we address. Two engineering constraints make it non-trivial. First, capability is multi-dimensional; a single model can be strong at code while weak at planning, so global ranking is misleading. Second, both cost and latency vary across models by an order of magnitude, so the wrong choice penalizes either accuracy or the bill.

\textbf{The main idea of this paper} is that routing across whole pretrained models can be decided in a compact six-dimensional capability space, in which both the query (what it demands) and the model (what it offers) are vectors, and the cheapest model that geometrically covers the query is the right answer.

This paper contributes:
\begin{enumerate}[noitemsep,topsep=3pt,leftmargin=1.4em,label=\arabic*.]
  \item We frame whole-model routing across a heterogeneous LLM pool as the \emph{Mixture-of-Models} (MoM) paradigm and show that, on Dataset~A, MoM via Brick recovers $\mathbf{24\%}$ of the achievable headroom between the best single model and the three-model oracle, with the residual $76\%$ of the gap reflecting information the router does not yet extract (§\ref{sec:mom}, §\ref{sec:results}).
  \item We show that two standard superficial routing strategies (domain-based and length/keyword-based) fail to capture intra-domain difficulty and underperform always-\texttt{kimi} on Dataset~A (§\ref{sec:superficial}, §\ref{sec:baselines}).
  \item \textbf{Brick}, a calibrated MoM router with an explicit capability-vector representation and a user-facing preference knob $r\!\in\![-1,1]$ that exposes the quality-vs-spend trade-off as a deploy-time parameter (§\ref{sec:brick}).
  \item Experimental results on Dataset~A showing Brick exceeds the best single-model baseline by $+1.96$\,pp in accuracy at lower cost and less than half its median end-to-end latency (§\ref{sec:results}, §\ref{sec:latency}).
\end{enumerate}

\section{The Mixture-of-Models (MoM) Paradigm}\label{sec:mom}
Training a new frontier model from scratch costs tens of millions of dollars and months of compute. A simpler question is: why not use the models already available? The market offers a wide range of pretrained models at different price points and capability profiles. Rather than collapsing this diversity into a single choice, we decided to build a custom architecture on top of it. We call this the \emph{Mixture of Models} (MoM) paradigm: a pool of whole pretrained models, dispatched one per query through an external router. The unit of routing is a model, not a layer. MoM differs from \emph{Mixture of Experts} (MoE), where sparse sub-networks inside a single model are activated per token; here the mixing happens at deployment time, across independent systems.

MoM is appealing for three operational reasons:

\begin{itemize}\setlength{\itemsep}{3pt}
  \item \textbf{Heterogeneous pricing.} Frontier commercial models charge ten to one hundred times what local open-weight models cost. A router that uses them only when needed makes the median query cheap without losing the long tail.
  \item \textbf{OW+CW bridging.} An MoM pool can mix open-weight models hosted in a controlled environment with closed-weight commercial APIs. The router lets an operator capture the data-locality and price benefits of open weights on easy queries while reserving expensive closed-weight calls for hard ones.
  \item \textbf{Capability complementarity.} Models from different vendors and training recipes often have complementary error sets. The three-model oracle on Dataset~A reaches $83.25\%$ (Table~\ref{tab:standalone_response_accuracy}), well above any single model; the mechanism behind this complementarity (refusal patterns, per-capability error sets) is analyzed in §\ref{sec:results}.
\end{itemize}

The rest of this paper takes Brick as a concrete instantiation of MoM. The geometric routing rule and the calibration procedure are properties of Brick, not of the MoM paradigm itself; alternative MoM realizations are possible and are discussed in §\ref{sec:discussion}.

\section{Limitations of Superficial Routing}\label{sec:superficial}
Two routing strategies recur in production systems and deserve to be ruled out before introducing Brick.

\textbf{Domain-based routing} maps the query to a coarse domain (science, math, humanities, code) and dispatches to a per-domain best model. The hidden assumption is that, conditional on the domain, models rank uniformly. This assumption fails because difficulty varies sharply inside a domain: a request to write a Python calculator and a request to solve a Riemann-zeta question are both ``code'' or both ``math'' depending on framing, but require different models. Domain-based routing collapses this variance and over-pays on easy queries.

\textbf{Length- and keyword-based routing} maps the query to a complexity score from token count and regex hits. The assumption is that long or jargon-rich queries are hard, and short queries are easy. This also fails: a one-line question may target an unsolved problem, while a long boilerplate prompt may wrap a trivial transformation. Token count is a poor proxy for required capability.

The remainder of this paper treats routing as a decision in capability space, not in surface-feature space.

\section{Model Pool}\label{sec:pool}
We use a fixed three-model pool:
\[
  \mathcal{M} =
  \{\texttt{qwen3.5-9b},\ \texttt{deepseek-v4-flash},\ \texttt{kimi2.6}\}.
\]
The pool spans three quality-and-cost regimes: \texttt{qwen3.5-9b} as a state-of-the-art under-$10$B-parameter open-weight model, \texttt{deepseek-v4-flash} as a state-of-the-art under-$500$B open-weight model, and \texttt{kimi2.6} as a frontier open-source contender.

\paragraph{Two cost objects.} The paper distinguishes a dimensionless routing-math scalar $c_m$ from the realised per-call dollar cost $a_m^{\$}$:
\begin{itemize}\setlength{\itemsep}{2pt}
\item $c_m$ is the input to Brick's cost-penalty term $\beta\,c_m$ inside the routing math of §\ref{sec:brick-math}. It is locked at calibration time and treated as a design-time constant on the same scale as the routing-distance $D_m$ (which lives in logit space, $O(1)$). For our three-model pool we use
\[
  c_{\text{qwen}}{=}0.10,\quad c_{\text{ds4}}{=}0.40,\quad c_{\text{kimi}}{=}0.60,
\]
chosen to match the per-1M-output-token price ratio at the time of evaluation under a normalization $p_{\mathrm{ref}}{=}1.00\ \text{USD}/10^6\ \text{tokens}$.
\item $a_m^{\$}$ is the real per-call dollar cost an operator would pay on Dataset~A, integrated over the observed token distribution of each model under public list prices:
\[
  a_m^{\$} = \frac{1}{N}\sum_{i=1}^{N} \frac{t^{\text{in}}_{i,m}\, p^{\text{in}}_m + (t^{\text{comp}}_{i,m} + t^{\text{reas}}_{i,m})\, p^{\text{out}}_m}{10^6}.
\]
At evaluation time the OpenRouter listed prices, in USD per $10^6$ tokens, were
\begin{align*}
  (p^{\text{in}}_{\text{qwen}},\, p^{\text{out}}_{\text{qwen}}) &= (0.04,\, 0.15), \\
  (p^{\text{in}}_{\text{ds4}},\, p^{\text{out}}_{\text{ds4}}) &= (0.20,\, 0.40), \\
  (p^{\text{in}}_{\text{kimi}},\, p^{\text{out}}_{\text{kimi}}) &= (0.73,\, 3.49),
\end{align*}
yielding $a_{\text{qwen}}^{\$}{=}\$0.001386$, $a_{\text{ds4}}^{\$}{=}\$0.002895$, $a_{\text{kimi}}^{\$}{=}\$0.030703$ per call (reasoning tokens are billed at the output rate by every vendor in our pool). The \texttt{deepseek-v4-flash} prices are the launch snapshot (released 2026-04-24; DeepSeek subsequently cut the cache-miss input to $\$0.14$ and the output to $\$0.28$ on 2026-04-26, and OpenRouter later dropped these to $\$0.10/\$0.20$). The relative spread $a_{\text{kimi}}^{\$}/a_{\text{ds4}}^{\$}\approx 10.6{\times}$ is wider than a simple per-1M-token price ratio would predict, and varies between $7{\times}$ and $44{\times}$ per protocol because \texttt{kimi2.6} emits a large reasoning-token tail on short-answer protocols.
\end{itemize}

All cost columns in the result tables (§\ref{sec:results}, §\ref{sec:baselines}, Table~\ref{tab:baselines_full}, Figure~\ref{fig:cost_pareto}, Table~\ref{tab:agent_step_blow_up}) report $a_m^{\$}$. The Brick routing math (§\ref{sec:brick-math}) uses $c_m$. The mapping between the two is fixed once at calibration time and is not updated at inference. The full per-protocol cost matrix is in \path{scientificv1/data/reports/cost_audit/hf_verbose_means.md}.

\section{Brick2 Dataset A}\label{sec:dataset}

\paragraph{Terminology.} Throughout the paper, a query is \emph{solved} if the model's response passes the per-protocol grader; otherwise it is \emph{unsolved}. The unsolved class aggregates four distinct behaviors: incorrect answer, refusal (``I don't know'' / empty response), format violation (IFEval/IFBench), and code that fails its unit tests. We reserve the word \emph{failure} for grader-level or unit-test-level failures specifically; at the dataset-aggregate level we use ``does not solve'' or ``unsolved''. The prevalence of refusals and their routing implications are discussed in §\ref{sec:discussion}; refusal is counted toward the unsolved class in every table.

Brick is evaluated on \textbf{Brick2 Dataset~A} (henceforth Dataset~A, distributed as \texttt{brick2-dataset-a-eval}), a stratified routing-evaluation benchmark of $N{=}5{,}504$ queries covering six capability dimensions. Table~\ref{tab:composition} summarizes the per-source composition. The corpus is built from fourteen license-clean upstream sources and three custom curated subsets. Each row carries a deterministic identifier, a typed evaluation payload, pre-computed token counts under the three target tokenizers, and a license tag enabling compliant redistribution.

\begin{table*}[!t]
\centering
\small
\caption{Per-source composition of Dataset~A ($N{=}5{,}504$). Counts may differ from the original release ($N{=}5{,}339$) due to the addition of $165$ planning multi-turn rows in revision $v0.4$.}
\label{tab:composition}
\begin{tabular}{l l r l}
\toprule
\textbf{Dimension} & \textbf{Source} & \textbf{Count} & \textbf{License} \\
\midrule
\multirow{2}{*}{instruction\_following}
  & IFEval~\cite{zhou2023ifeval}                & 541       & apache-2.0     \\
  & IFBench~\cite{pyatkin2025ifbench}           & 300       & ODC-BY-1.0     \\
\midrule
coding
  & LiveCodeBench-v6~\cite{jain2024livecodebench} & $1{,}000$ & cc-by-4.0      \\
\midrule
\multirow{3}{*}{math\_reasoning}
  & MATH-500~\cite{hendrycks2021math}            & 500       & mit            \\
  & AIME-2025                                    & 30        & qwen-derived   \\
  & GSM8K~\cite{cobbe2021gsm8k}                  & 470       & mit            \\
\midrule
\multirow{3}{*}{world\_knowledge}
  & SimpleQA~\cite{wei2024simpleqa}              & 700       & mit            \\
  & MMLU-Pro-Humanities~\cite{wang2024mmlupro}   & 102       & mit            \\
  & GPQA-Diamond~\cite{rein2024gpqa}             & 0 (gated) & cc-by-4.0      \\
\midrule
\multirow{3}{*}{creative\_synthesis}
  & EQ-Bench-Creative-v3~\cite{paech2024eqbench} & 96        & EQ-Bench terms \\
  & LitBench~\cite{stanford2025litbench}         & 500       & qwen-derived   \\
  & Custom-Validated                             & 100       & qwen-derived   \\
\midrule
\multirow{3}{*}{planning\_agentic}
  & BFCL-v4~\cite{yan2024bfcl}                   & 500       & apache-2.0     \\
  & tau-bench~\cite{yao2024taubench}             & 165       & mit            \\
  & Planning-Custom (incl. multi-turn)           & 500       & qwen-derived   \\
\midrule
\multicolumn{2}{l}{\textbf{Total}} & $\mathbf{5{,}504}$ & \\
\bottomrule
\end{tabular}
\end{table*}

\subsection{Quality validation and oracle bound}
The construction pipeline is validated in three tiers: deterministic schema and length checks, LLM-as-judge calibration against an audited reference subset, and a three-reviewer manual audit on $n{=}102$ stratified samples. The three reviewers are the two authors (Francesco Massa, Marco Cristofanilli) and an LLM reviewer (Claude Sonnet 4.6) acting as an independent third rater under the same rubric.

\subsection{Grading methods}\label{sec:grading}
Dataset~A's ten scoring protocols split into two grading regimes that differ in the number of judge models involved, and a third \emph{variance-characterization} regime applied to a single protocol only. \textbf{Eight protocols are deterministic and use no judge model} (zero judges): coding (\texttt{lcb\_unit\_test}) is graded by the LiveCodeBench unit-test harness; \texttt{math\_reasoning} (\texttt{gsm8k\_final\_answer}, \texttt{math\_equiv}) by SymPy and regex matching against gold answers; \texttt{instruction\_following} (\texttt{ifeval\_constraint\_check}) by the IFEval and IFBench programmatic checkers; \texttt{mcq\_letter} by deterministic letter extraction; and \texttt{tool\_call\_match} (single- and multi-turn) by the BFCL AST/state checkers. \textbf{Two protocols use a single LLM judge} (one judge): \texttt{world\_knowledge}'s \texttt{llm\_judge\_factual} and the open-ended \texttt{rubric\_judge} (creative writing and agentic planning) are graded by \texttt{openai/gpt-5.4-mini} alone, cross-provider relative to the evaluated pool, at temperature~$0$. \textbf{One protocol additionally runs a three-judge panel} (three judges, used only for variance characterization, not for the headline accuracy numbers): the agentic planning \texttt{rubric\_judge} is also re-graded by a fixed three-judge panel (\texttt{openai/gpt-5.4-mini}, \texttt{mistralai/mistral-small-2603}, \texttt{z-ai/glm-5-turbo}), each from a family disjoint from the evaluated pool, with majority vote at temperature $0$; this panel exists to quantify judge-induced variance on the most subjective protocol and is reported in dataset~A's evaluation chapter, not used to compute the response-accuracy numbers in this paper. Throughout this paper, ``response accuracy'' for any model or router refers to the per-protocol grader of the originating protocol under the regime listed above.

Aggregate per-model response accuracy on Dataset~A is reported in Table~\ref{tab:standalone_response_accuracy}, together with the achievable three-model oracle bound. A query is counted as solved if the model's response passes the per-protocol grader. The oracle counts a query as solved if any of the three models would have solved it; the residual $922$ queries are unsolvable by the current pool.

\begin{table}[H]
\centering
\caption{Standalone per-model response accuracy and three-model oracle bound on Dataset~A.}
\label{tab:standalone_response_accuracy}
\small
\begin{tabular}{l r r}
\toprule
\textbf{System} & \textbf{Solved} & \textbf{Accuracy} \\
\midrule
always \mdlogo{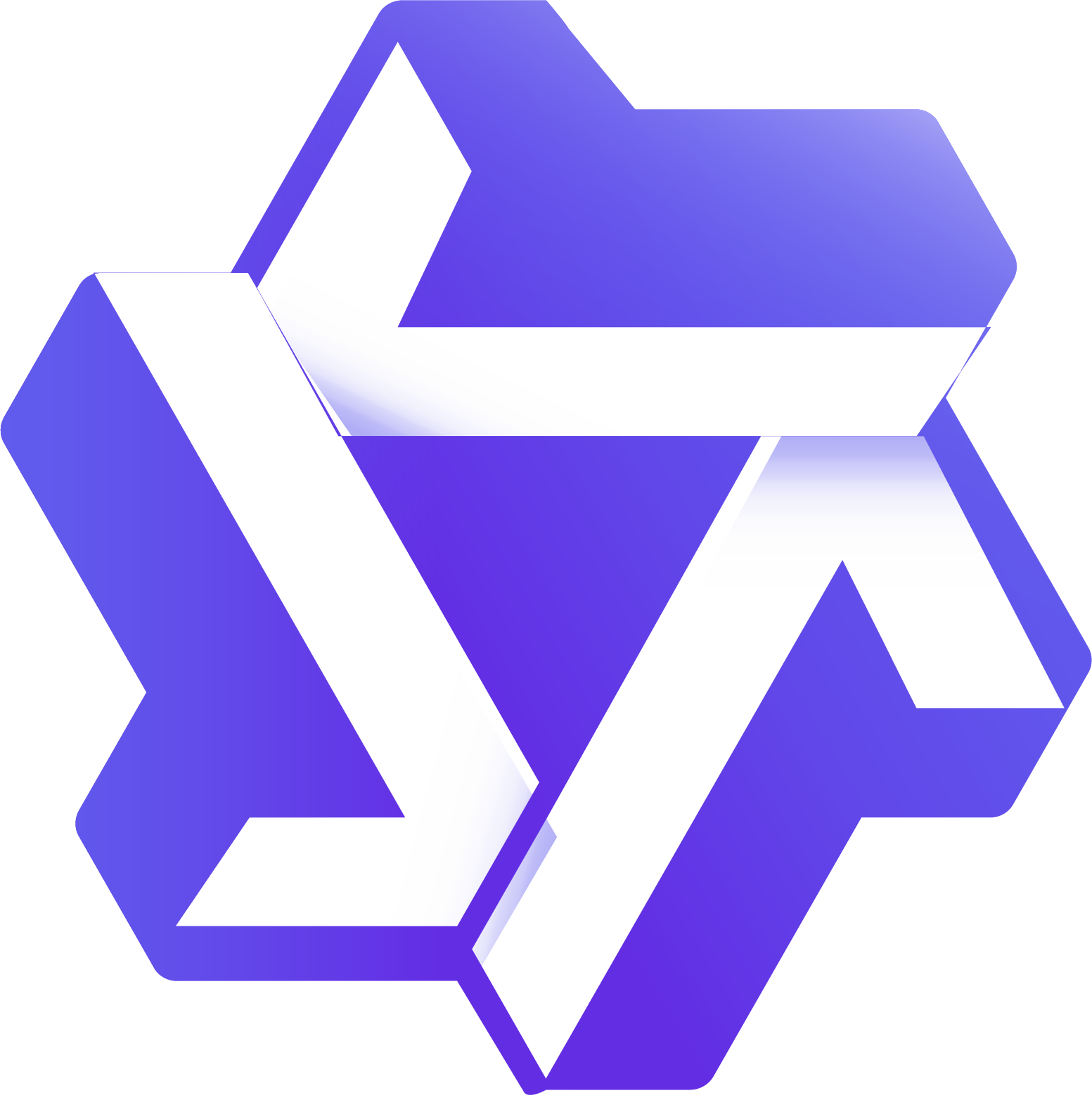}\,\texttt{qwen}    & $3{,}477 / 5{,}504$ & $63.17\%$ \\
always \mdlogo{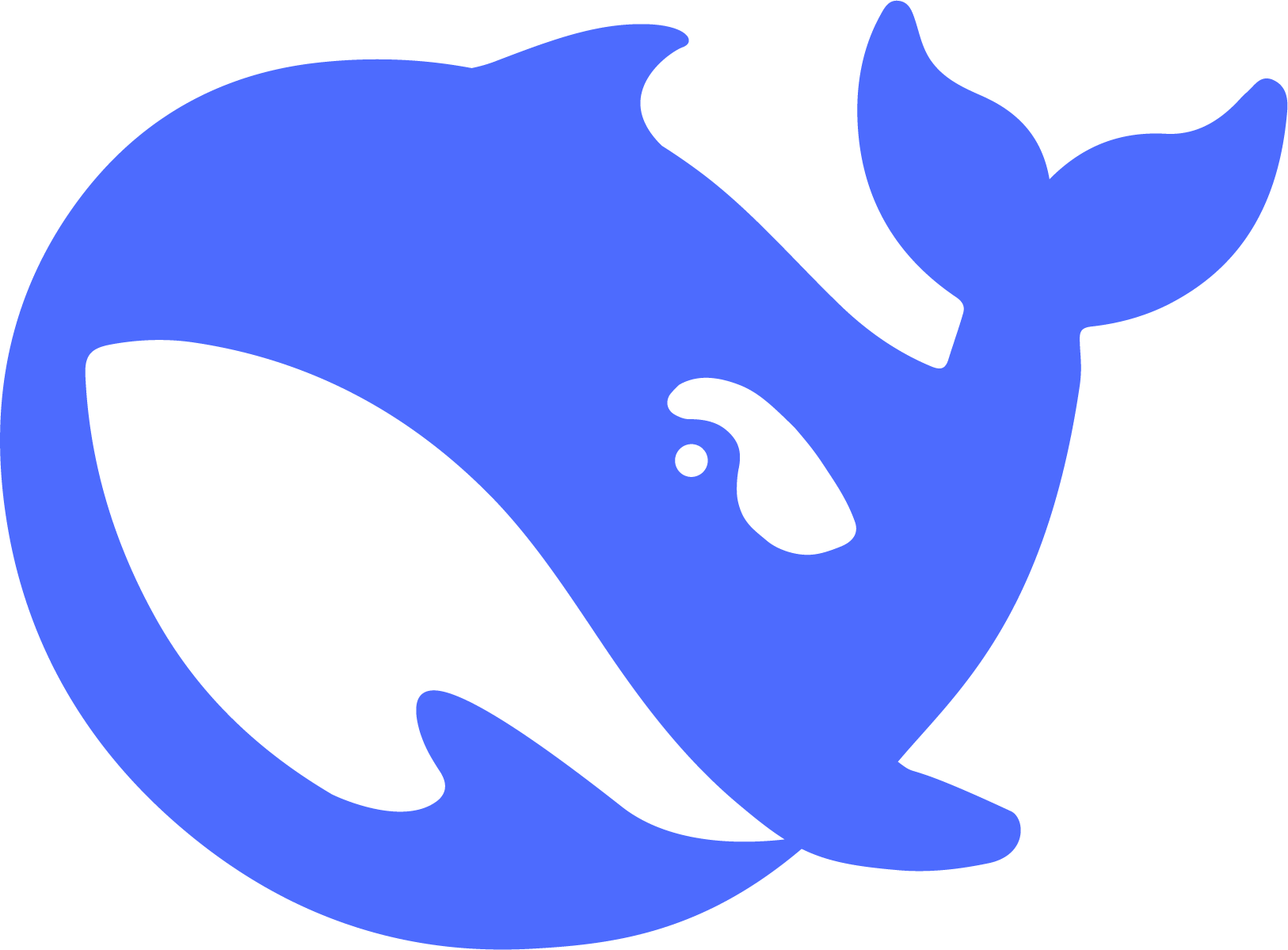}\,\texttt{ds4}      & $4{,}056 / 5{,}504$ & $73.69\%$ \\
always \mdlogo{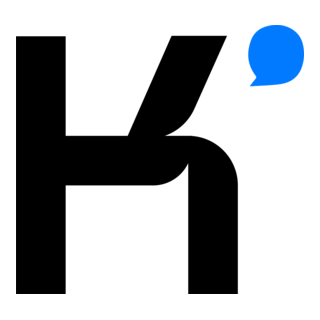}\,\texttt{kimi}    & $4{,}129 / 5{,}504$ & $75.02\%$ \\
\midrule
three-model oracle      & $4{,}582 / 5{,}504$ & $\mathbf{83.25\%}$ \\
unsolvable (any model)  & $922 / 5{,}504$    & $16.75\%$ \\
\bottomrule
\end{tabular}
\end{table}

Notably, \texttt{kimi2.6} is not the per-query oracle: it does not solve $1{,}375$ queries (incorrect, refused, or format-violating responses), of which $453$ are correctly answered by at least one cheaper model. The \emph{two-model} oracle restricted to \texttt{kimi}$+$\texttt{ds4} already reaches $81.96\%$ (a $+6.94$ pp gain over \texttt{kimi} alone), and the \emph{three-model} oracle on the full pool $\{$\texttt{qwen},\texttt{ds4},\texttt{kimi}$\}$ reaches $83.25\%$ as reported in the table above (a $+8.23$ pp ceiling over \texttt{kimi}). Only $16.75\%$ of queries remain unsolvable by any model in the pool. This residual headroom is the upside an MoM router can in principle capture.

The dataset (officially \texttt{brick2-dataset-a}) is publicly available at \href{https://huggingface.co/datasets/regolo/brick2-dataset-a-eval}{regolo/brick2-dataset-a-eval}.

\subsection{Evaluation protocol}\label{sec:eval-protocol}
All routers in this paper, both Brick and the external baselines of §\ref{sec:baselines}, are scored on the same Dataset~A test set under a single \emph{selected-answer accuracy} metric: a routed query counts as correct if and only if the model the router actually dispatches the query to solves it, as judged by Dataset~A's per-protocol grader. Per-model correctness ($\texttt{qwen\_correct}$, $\texttt{ds4\_correct}$, $\texttt{kimi\_correct}$) is read directly from the public Dataset~A \texttt{results} subset, so all systems share an identical ground-truth source with no router-specific labelling. Cost per query is the normalised cost vector $a_m$ of §\ref{sec:pool} evaluated at the dispatched model; for cascade-style baselines we additionally report a \emph{cumulative} cost that adds the cost of every earlier cascade stage paid before the finally accepted model, since those rejected calls also bill on a real deployment. No query is excluded, no router-specific re-grading is performed, and no fallback substitution is applied to the ground truth: when no model in the pool solves a query (the $922 / 5{,}504$ ``unsolvable'' rows of Table~\ref{tab:standalone_response_accuracy}), every router scores it as incorrect, including the oracle.

\section{External Router Baselines}\label{sec:baselines}
We benchmark four published routing strategies on Dataset~A under the same three-model pool, all scored under the selected-answer accuracy metric of §\ref{sec:eval-protocol}, with public reference checkpoints in zero-shot mode (no fitting on Dataset~A). The four checkpoints fall into two structural classes (a \emph{single-step} class that picks one model per query in one shot, and a \emph{cascade} class that probes models in cost order and escalates on rejection), so we organize the section accordingly.

\subsection{Single-step routers}\label{sec:baselines-singlestep}
RouteLLM~\cite{ong2024routellm} in its binary and tournament variants is the representative single-step router in our pool. Its pairwise-preference formulation is designed for two-model dispatch; on a three-model pool with our cost spread, the public checkpoint dispatches essentially $100\%$ of queries to \texttt{kimi}, matching always-\texttt{kimi}'s $75.02\%$ selected-answer accuracy at $\$0.030703$ per call (Table~\ref{tab:baselines_full}). The tournament variant adds a second pairwise call without changing the dispatched model on any row in our split. Binary preference routers do not extend cleanly to $K{\ge}3$ pools without recursive pairing, and the recursion would still produce a single-step decision per pair, which is what an MoM router already provides natively in one step.

\subsection{Cascade routers}\label{sec:baselines-cascade}
FrugalGPT-style cascade routing~\cite{chen2023frugalgpt} and Cascade Routing~\cite{jitkrittum2024cascade} are the two cascade-class baselines. They share the same structural pattern (probe the cheapest model, observe its answer, accept or escalate to the next-cheapest based on a correctness scorer) and differ only in the type of accept/reject scorer they install on top: FrugalGPT uses a fine-tuned DistilBERT scorer that predicts answer correctness from the (query, response) pair, with a fixed per-stage threshold, while Cascade Routing uses a probabilistic per-model utility scorer (a calibrated logistic regression on per-query features) with an adaptive threshold derived from a cost-aware utility. Both consequently probe models in cost order (\texttt{qwen}\,$\to$\,\texttt{ds4}\,$\to$\,\texttt{kimi}) and accept the first model whose scorer clears the threshold. On Dataset~A, FrugalGPT's gates escalate roughly two thirds of queries to \texttt{ds4} and only a residual $3\%$ to \texttt{kimi}, reaching $69.42\%$ selected-answer accuracy at a per-call cost of $\$0.002800$ (or $\$0.004114$ once the rejected-cascade calls are amortized in: $\approx 1.66\times$ cost overhead relative to the accepted-model cost alone). Cascade Routing spreads load across all three tiers ($18 / 61 / 21$) and reaches $73.40\%$ selected-answer accuracy at $\$0.006113$ per call (the public checkpoint we evaluate scores all three models in parallel rather than probing sequentially, so cumulative equals per-call cost; a strictly sequential variant of the same scorer would add the rejected-stage costs of the qwen and ds4 calls, totaling $\approx \$0.008746$). Both accuracies are below the $75.02\%$ of always-\texttt{kimi}, despite the cascade structure paying for additional sequential calls on a large fraction of queries.

\subsection{Pure routing vs cascade in an agentic regime}\label{sec:baselines-vs-cascade}
\textbf{Cascade routing is structurally wasteful for agentic deployments.} The single-call overheads above ($1.0{-}1.47\times$ on the public checkpoints, with a strictly sequential variant of Cascade Routing reaching $1.43\times$) already understate the problem: they assume the router is invoked once per user request. In agentic deployments the assumption breaks. A single user request fans out into $N$ LLM calls (tool-use loops, planner-executor chains, multi-step reasoning over intermediate observations), and the router is invoked on every one of them. A pure single-step router pays a fixed per-call overhead (the routing decision itself) on each of the $N$ calls. A cascade router additionally pays the rejected-cascade cost on each of the $N$ calls, because every step independently re-runs the cascade gate (a cascade does not remember that step $k{-}1$ already failed the cheap model; the request at step $k$ is a new query and starts the cascade from the cheapest tier again).

Concretely, with our measured per-decision overheads, the bill for an $N$-step agent trajectory at the max profile scales as in Table~\ref{tab:agent_step_blow_up}. By $N{=}10$ the cascade routers are paying $1.0{-}1.47\times$ the bill of an equivalently routed pure single-step pipeline on top of the underlying inference cost, and the gap grows linearly with $N$. There is no threshold tuning that makes this overhead go away: it is the structural cost of probing models sequentially, paid every time the agent decides to call a model. Latency suffers the same fate, because cascade stages are sequential by construction (each gate must observe the rejected model's output before it can fire), so each rejected stage adds a full round-trip to the user's critical path. For long-horizon agents this is the difference between a few seconds and tens of seconds of perceived idle time per request, on top of the bill.

\begin{table*}[!t]
\centering
\caption{Cumulative cost of an $N$-step agent trajectory at the max-quality operating point. Pure routers (Brick max, RouteLLM, always-\texttt{kimi}) pay the cost of the dispatched model on each step. Cascade routers (FrugalGPT, Cascade Routing) additionally pay the cost of every rejected cascade stage on each step. All numbers are USD per agent trajectory, consistent with Table~\ref{tab:baselines_full}.}
\label{tab:agent_step_blow_up}
\small
\begin{tabular}{l r r r r}
\toprule
\textbf{Router} & $N{=}1$ & $N{=}5$ & $N{=}10$ & $N{=}20$ \\
\midrule
Brick max          & $\$0.022083$ & $\$0.110415$ & $\$0.220830$ & $\$0.441660$ \\
RouteLLM (binary or tournament) & $\$0.030703$ & $\$0.153515$ & $\$0.307030$ & $\$0.614060$ \\
always-\texttt{kimi} (no router) & $\$0.030703$ & $\$0.153515$ & $\$0.307030$ & $\$0.614060$ \\
\midrule
FrugalGPT cascade  & $\$0.004114$ & $\$0.020570$ & $\$0.041140$ & $\$0.082280$ \\
Cascade Routing (parallel scorer)   & $\$0.006113$ & $\$0.030565$ & $\$0.061130$ & $\$0.122260$ \\
\bottomrule
\end{tabular}
\end{table*}

FrugalGPT remains cheaper in absolute terms here because it accepts \texttt{qwen}'s answer on roughly $33\%$ of queries (its self-verification gate fires early), but it pays for that cheap dispatch with $-7.56$\,pp of selected-answer accuracy versus Brick max and $-3.98$\,pp versus Cascade Routing (Table~\ref{tab:baselines_full}). The point of Table~\ref{tab:agent_step_blow_up} is not that one cascade is cheaper than another, but that the cascade family pays a per-step structural overhead on every step of the agent loop, while a pure router pays the routing decision once per step and forwards a single LLM call. For agentic workloads this is the difference that matters, and it is paid regardless of how well the cascade thresholds are tuned.

\begin{table*}[!t]
\centering
\caption{Unified baselines table on Dataset~A: single-model baselines, external routers, and Brick MoM profiles. Selected-answer accuracy counts a routed query correct iff the dispatched model solves it (deployment-facing). Route-exact accuracy counts it correct iff the routed model matches the cheapest model that actually solves the query (cost-efficiency-facing, fallback to \texttt{kimi} on unsolvable rows). Average cost is the normalized per-call USD cost of the model invoked; for cascade-style baselines we also report the \emph{cumulative} cost, which adds the cost of any earlier cascade stage paid before acceptance. Distribution reports the share of queries finally routed to each model (qwen / ds4 / kimi).}
\label{tab:baselines_full}
\small
\setlength{\tabcolsep}{4pt}
\begin{tabular}{l r r r r c}
\toprule
\textbf{Router} & \textbf{Sel.-ans.\,acc.} & \textbf{Route-exact acc.} & \textbf{Avg.\,cost (USD)} & \textbf{Cum.\,cost (USD)} & \textbf{Dist.\,q\,/\,ds4\,/\,kimi (\%)} \\
\midrule
always \mdlogo{qwen}\,qwen & $63.17\%$ & $63.17\%$ & $\$0.001386$ & n/a & $100 / 0 / 0$ \\
always \mdlogo{ds4}\,ds4   & $73.69\%$ & $15.55\%$ & $\$0.002895$ & n/a & $0 / 100 / 0$ \\
always \mdlogo{kimi}\,kimi & $75.02\%$ & $21.28\%$ & $\$0.030703$ & n/a & $0 / 0 / 100$ \\
\midrule
RouteLLM binary      & $75.02\%$ & $21.31\%$ & $\$0.030702$ & n/a          & $0 / 0 / 100$ \\
RouteLLM tournament  & $75.02\%$ & $21.31\%$ & $\$0.030702$ & n/a          & $0 / 0 / 100$ \\
FrugalGPT cascade    & $69.42\%$ & $31.03\%$ & $\$0.002800$ & $\$0.004114$ & $33 / 64 / 3$ \\
Cascade Routing      & $73.40\%$ & $28.96\%$ & $\$0.006113$ & $\$0.006113$ & $18 / 61 / 21$ \\
\midrule
Brick min ($r{=}{-}1.0$)     & $63.17\%$            & $\mathbf{63.17\%}$ & $\$0.001386$ & n/a & $100 / 0 / 0$ \\
Brick low ($r{=}{-}0.5$)     & $71.62\%$            & $48.66\%$          & $\$0.003557$ & n/a & $45 / 51 / 5$ \\
Brick neutral ($r{=}0$)      & $74.11\%$            & $40.35\%$          & $\$0.006513$ & n/a & $29 / 56 / 15$ \\
Brick high ($r{=}{+}0.5$)    & $76.24\%$            & $35.63\%$          & $\$0.014905$ & n/a & $21 / 35 / 44$ \\
Brick max ($r{=}{+}1.0$)     & $\mathbf{76.98\%}$   & $17.77\%$          & $\$0.022083$ & n/a & $0 / 31 / 69$ \\
\midrule
3-model oracle       & $\mathbf{83.25\%}$ & $\mathbf{100.00\%}$ & n/a & n/a & n/a \\
\bottomrule
\end{tabular}
\end{table*}

\subsection{Cost vs performance frontier}\label{sec:cost}
\begin{figure*}[!t]
\centering
\includegraphics[width=\textwidth]{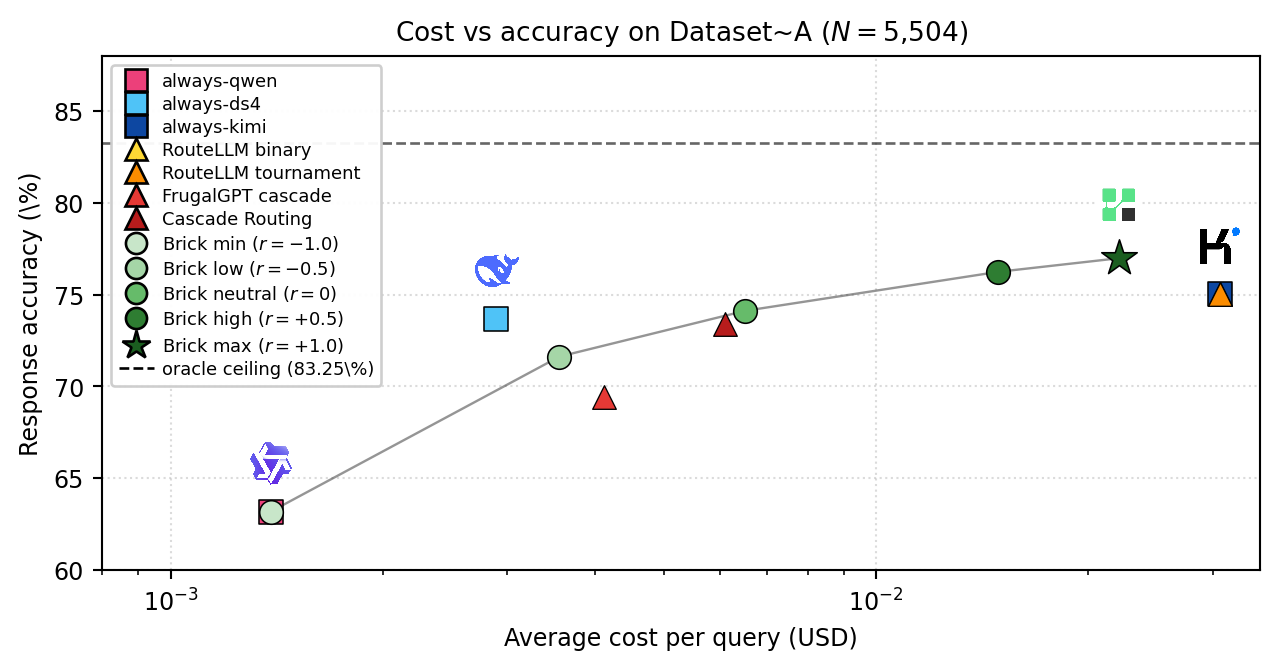}
\caption{Cost vs response accuracy on Dataset~A. Single-model baselines are squares, external routers are triangles, Brick (MoM) profiles are circles (max profile is a star). The dashed line marks the three-model oracle ceiling.}
\label{fig:cost_pareto}
\end{figure*}

Figure~\ref{fig:cost_pareto} plots response accuracy against average USD cost per query for the three single-model baselines, the four external routers, the five Brick preference profiles introduced in §\ref{sec:brick}, and the three-model oracle ceiling. Brick's max-quality profile ($76.98\%$ at $\$0.022083$) dominates always-\texttt{kimi} ($75.02\%$ at $\$0.030703$) on both axes: higher accuracy at $\mathbf{28\%}$ lower per-call cost ($1.39{\times}$ cheaper). The Brick low profile reaches $71.62\%$ at $\$0.003557$, beating FrugalGPT's $69.42\%$ accuracy by $+2.20$\,pp, achieved by preferring \texttt{deepseek-v4-flash} ($51\%$ dispatch share at this profile) over \texttt{kimi2.6} (effectively zero, $5\%$) on the queries that \texttt{qwen} cannot solve. The Brick neutral profile reaches $\mathbf{74.11\%}$ at $\$0.006513$, $\mathbf{4.71{\times}}$ cheaper than always-\texttt{kimi} for only a $0.91$\,pp accuracy give-up, and beats Cascade Routing's $73.40\%$ by $+0.71$\,pp at comparable cost; it dispatches $56\%$ to \texttt{ds4} and only $15\%$ to \texttt{kimi}, leveraging the \texttt{ds4} sweet-spot on the \texttt{world\_knowledge} capability where \texttt{ds4} actually outperforms \texttt{kimi2.6} ($49\%$ vs $34\%$ correctness). Brick high ($76.24\%$ at $\$0.014905$) sits $+2.84$\,pp above Cascade Routing. The Brick curve traces a Pareto front above the external routers across the operating range, with the largest gap in the mid-to-high cost region as the curve climbs toward the three-model oracle ceiling at $83.25\%$. The remainder of the paper unpacks how Brick recovers this headroom and how much of the oracle ceiling stays out of reach.

\subsection{Routing efficiency vs response quality}\label{sec:two-metrics}

The selected-answer accuracy used so far answers a deployment question (``did the user get a correct answer?''), but it is not the only metric of interest for a router. A complementary metric is \emph{route-exact accuracy}, which answers a cost-efficiency question (``did the router avoid paying for an expensive model when a cheaper one would have solved the query?''). The two are formally defined on the same per-model correctness flags $z_{i,m}\!\in\!\{0,1\}$ used in §\ref{sec:eval-protocol}:
\setlength{\abovedisplayskip}{10pt}\setlength{\belowdisplayskip}{10pt}%
\[
\begin{aligned}
  \mathrm{SelectedAns} &= \tfrac{1}{N}{\textstyle\sum\nolimits_i}\, z_{i, m_i^\star}, \\[-1pt]
  \mathrm{RouteExact}  &= \tfrac{1}{N}{\textstyle\sum\nolimits_i}\, \mathbf{1}[m_i^\star{=}g_i].
\end{aligned}
\]
where $g_i = \arg\min_{m: z_{i,m}=1} a_m$ is the cheapest correct model on query $i$ (with fallback to the strongest model when no model in the pool solves it; this matches the calibration objective used during the Bayesian sweep, §\ref{sec:brick-knob}). Both are evaluated on the same $5{,}504$ queries with the same per-protocol graders.

The two metrics are deliberately not aligned and a router cannot game both with a single fixed strategy. A router that always picks \texttt{kimi} maximises selected-answer ($75.02\%$) but collapses route-exact ($21.28\%$), because it overpays for the $63\%$ of queries that \texttt{qwen} would already have solved. A router that always picks \texttt{qwen} hits route-exact $63.17\%$ (every cheapest-correct row where \texttt{qwen} is the cheapest correct option) but loses on selected-answer ($63.17\%$, since it fails every row that requires \texttt{ds4} or \texttt{kimi}). Cascade routers fall in between, but pay both metrics: they pick the wrong model relative to the cheapest-correct label, \emph{and} they pay the rejected cascade stage for queries that the first stage failed to solve.

Table~\ref{tab:baselines_full} reports both metrics for every system on Dataset~A. Three facts stand out. First, \textbf{Brick's preference knob traces the entire frontier between the two metrics}: setting $r{=}{-}1$ collapses the router to always-\texttt{qwen} behaviour and recovers $63.17\%$ on both metrics, while setting $r{=}{+}1$ optimises for response quality (max-profile $76.98\%$ selected-answer) at the explicit cost of route-exact collapsing to $17.79\%$. The latter is not a regression: the max profile dispatches $69\%$ of queries to \texttt{kimi}, knowingly over-routing on the $63\%$ that \texttt{qwen} could already solve, because the operator has signalled (via $r{=}{+}1$) that the bound on response quality matters more than the cheapest-correct match. Second, \textbf{Brick low dominates FrugalGPT on response accuracy at competitive cost}: $71.62\%$ selected-answer ($+2.20$\,pp over FrugalGPT's $69.42\%$) and $48.66\%$ route-exact, above both Cascade Routing's $28.96\%$ and FrugalGPT's $31.03\%$. The mechanism is the routing preference for \texttt{deepseek-v4-flash} on queries that exceed \texttt{qwen}'s reach: at $r{=}{-}0.5$ Brick dispatches $51\%$ to \texttt{ds4} and only $5\%$ to \texttt{kimi}. Third, \textbf{Brick neutral beats Cascade Routing on accuracy at comparable cost}: $74.11\%$ selected-answer ($+0.71$\,pp over Cascade Routing's $73.40\%$) and $40.35\%$ route-exact, above both Cascade Routing's $28.96\%$ and FrugalGPT's $31.03\%$, with $56\%$ \texttt{ds4} dispatch absorbing the under-routing failures that an aggressive cheap-only baseline would miss.

\section{Brick: A Calibrated MoM Router}\label{sec:brick}

\subsection{Pipeline}\label{sec:brick-pipeline}
\begin{samepage}
Brick processes each query through a deterministic six-step pipeline:
\nopagebreak
\begin{enumerate}\setlength{\itemsep}{3pt}\setlength{\parskip}{0pt}
  \item trim leading and trailing whitespace and truncate the query to a $512$-rune budget \emph{for the classification stage only} (the full untruncated query is forwarded to the selected backend);
  \item match the query against high-precision keyword rules (e.g.\ explicit code fences) that can bias the capability distribution;
  \item compute a capability probability vector $p(x)\in\Delta^{D-1}$ (i.e.\ $p_c\!\ge\!0$ and $\sum_c p_c{=}1$) over a six-dimensional basis using a fine-tuned ModernBERT classifier;
  \item compute a complexity label and confidence using a separate complexity classifier, mapping to a base difficulty level $\tau_q$;
  \item compute a per-model capability score $J_m$ using the routing math of §\ref{sec:brick-math};
  \item select $m^{\star}=\arg\min_m J_m$ with deterministic tie-breaking.
\end{enumerate}
\end{samepage}

\paragraph{Routing-time preprocessing budget.} The $512$-rune truncation applies \emph{only} to the text handed to the capability and complexity classifiers (Step~1 above); the full query, including any long context, system prompt, or attached documents, is forwarded to the selected backend without modification. The budget is sized to be large enough for the two classifiers to identify the task intent from the opening of the query (system prompt, leading question, first lines of code or document), but small enough to keep routing latency negligible by avoiding the cost of tokenising and embedding long contexts purely for a routing decision. The budget is also aligned with the ModernBERT capability classifier's training sequence length ($512$ tokens), so $512$ runes is a conservative upper bound that guarantees the classifier sees an in-distribution input regardless of language or tokeniser. Inside the classifier stage, the ModernBERT output probabilities are renormalised to a probability simplex via $L_1$ normalisation with non-negative and \texttt{NaN}-safe handling (uniform fallback if the sum is zero), and the complexity classifier additionally truncates its tokenised input to $1024$ tokens at the tokeniser level. These choices are design-time constants (not part of the calibration search of §\ref{sec:brick-knob}) and live in \texttt{brickrouting/router.go} and \texttt{brickrouting/math.go}.

\begin{figure*}[!t]
\centering
\includegraphics{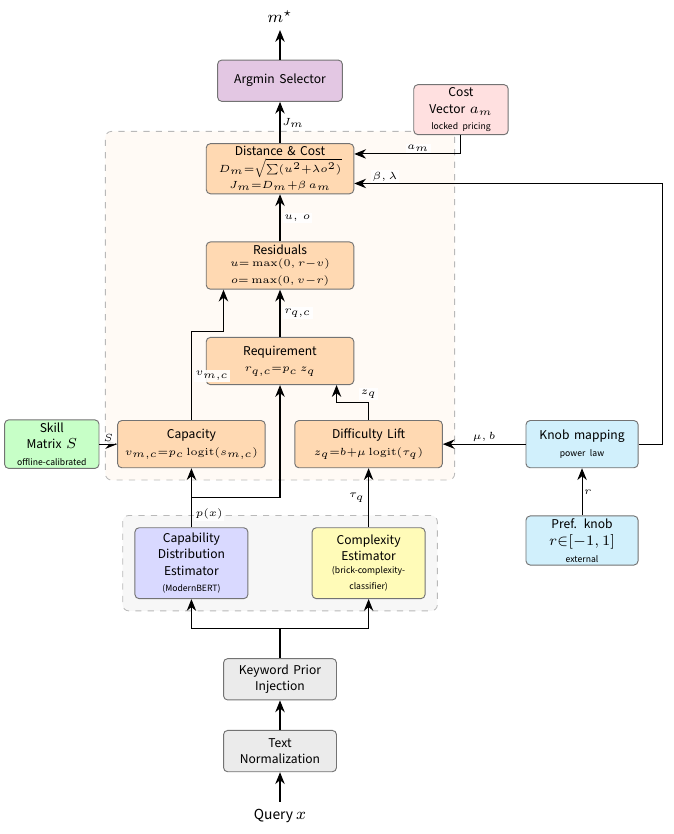}
\caption{Brick architecture. The query $x$ flows through text normalization and keyword prior injection, then branches into two parallel estimators: a ModernBERT capability distribution head producing $p(x)\!\in\!\Delta^{D-1}$, and a complexity head producing the difficulty target $\tau_q$. The routing math is decomposed into five sub-blocks: difficulty lift (raising $\tau_q$ to logit space $z_q$), per-capability requirement $r_{q,c}$ and per-model capacity $v_{m,c}$ builders, asymmetric under/over-capacity residuals $(u_{m,c}, o_{m,c})$, and the cost-penalized distance $J_m$. Three side inputs feed the math: the offline-calibrated skill matrix $S\!\in\!(0,1)^{K\!\times\!D}$ enters the capacity builder, the locked cost vector $a_m$ enters the distance \& cost combiner, and the external user-preference knob $r\!\in\![-1,1]$ is mapped via a calibrated power law to the four routing scalars $(\mu, b)$ feeding the difficulty lift and $(\beta, \lambda)$ feeding the distance \& cost block. The argmin selector returns $m^{\star}$.}
\label{fig:brick-arch}
\end{figure*}

\subsection{Worked example as motivation}\label{sec:brick-example}
Before the formal routing math, we show one full dispatch on a real Dataset~A query, computing every quantity numerically. The four routing scalars $(\mu, b, \beta, \lambda)$ used here are the neutral-profile defaults; they are functions of the preference knob $r\!\in\![-1,1]$ formally introduced in §\ref{sec:brick-knob} (Table~\ref{tab:brick-knob-defaults}). The skill matrix $S$ is the offline-calibrated per-model per-capability success table whose construction is detailed in §\ref{sec:brick-math}.


\textbf{Query.} \texttt{q\_03563} (dim: \texttt{planning\_agentic}): ``Can you help me with a few tasks? First, I am playing a game where I need to calculate the evolutionary fitness of a creature. ...''

\vspace{4pt}\noindent\textbf{Step 0. Preference knob.} The caller sets $r{=}0$ (\texttt{balanced} profile), so $p^{+}{=}p^{-}{=}0$ and the four routing scalars collapse to their locked production base values (Table~\ref{tab:brick-knob-defaults}):
\[
(\mu, b, \beta, \lambda) = (\mu_0, b_0, \beta_0, \lambda_0) = (1.07, 0.15, 0.63, 0.35).
\]

\vspace{4pt}\noindent\textbf{Step 1. Capability distribution} $p(x)$ from ModernBERT (six entries, sum to one):
\vspace{-2pt}\begin{center}\small
\begin{tabular}{l r r r r r r }
\toprule
$c$ & \texttt{cod} & \texttt{crea} & \texttt{ifo} & \texttt{math} & \texttt{plan} & \texttt{wld} \\
$p_c$ & 0.09 & 0.53 & 0.09 & 0.09 & 0.09 & 0.09 \\
\bottomrule
\end{tabular}\end{center}
\vspace{-4pt}{\footnotesize\noindent(Values rounded to two decimals for display; full-precision probabilities sum to one.)}

\vspace{4pt}\noindent\textbf{Step 2. Complexity and difficulty lift.} The complexity head labels the query \texttt{hard} with confidence $0.51$, blending into $\tau_q{=}0.802$. The effective difficulty logit is
\[
z_q \;=\; b + \mu\,\mathrm{logit}(\tau_q) \;=\; 0.15 + 1.07\cdot1.401 \;=\; 1.649.
\]

\vspace{4pt}\noindent\textbf{Step 3. Per-capability requirement} $r_{q,c}{=}p_c\,z_q$ (with $z_q{=}1.649$):
\vspace{-2pt}\begin{center}\small
\begin{tabular}{l r r r r r r }
\toprule
$c$ & \texttt{cod} & \texttt{crea} & \texttt{ifo} & \texttt{math} & \texttt{plan} & \texttt{wld} \\
\midrule
$p_c$ & 0.09 & 0.53 & 0.09 & 0.09 & 0.09 & 0.09 \\
$r_{q,c}$ & 0.16 & 0.87 & 0.16 & 0.16 & 0.16 & 0.16 \\
\bottomrule
\end{tabular}\end{center}
\noindent\textbf{Step 4. Per-model capacity} $v_{m,c}{=}p_c\,\mathrm{logit}(s_{m,c})$, with $S$ the offline-calibrated skill matrix:
\vspace{-2pt}\begin{center}\scriptsize
\begin{tabular}{l r r r r r r }
\toprule
$c$ & \texttt{cod} & \texttt{crea} & \texttt{ifo} & \texttt{math} & \texttt{plan} & \texttt{wld} \\
\midrule
$s_{qwen,c}$ & 0.715 & 0.512 & 0.810 & 0.912 & 0.577 & 0.180 \\
$v_{qwen,c}$ & 0.09 & 0.02 & 0.14 & 0.22 & 0.03 & -0.14 \\
$s_{ds4,c}$ & 0.821 & 0.658 & 0.863 & 0.935 & 0.621 & 0.489 \\
$v_{ds4,c}$ & 0.14 & 0.34 & 0.17 & 0.25 & 0.05 & -0.00 \\
$s_{kimi,c}$ & 0.904 & 0.752 & 0.870 & 0.944 & 0.642 & 0.344 \\
$v_{kimi,c}$ & 0.21 & 0.58 & 0.18 & 0.27 & 0.06 & -0.06 \\
\bottomrule
\end{tabular}\end{center}
\noindent\textbf{Step 5. Asymmetric residuals.} $u_{m,c}{=}\max(0, r_{q,c}{-}v_{m,c})$ (under-capacity, model too weak), $o_{m,c}{=}\max(0, v_{m,c}{-}r_{q,c})$ (over-capacity, overkill):
\vspace{-2pt}\begin{center}\scriptsize
\begin{tabular}{l r r r r r r }
\toprule
$c$ & \texttt{cod} & \texttt{crea} & \texttt{ifo} & \texttt{math} & \texttt{plan} & \texttt{wld} \\
\midrule
$u_{qwen,c}$ & 0.07 & 0.84 & 0.02 & 0.00 & 0.13 & 0.30 \\
$o_{qwen,c}$ & 0.00 & 0.00 & 0.00 & 0.07 & 0.00 & 0.00 \\
$u_{ds4,c}$ & 0.01 & 0.52 & 0.00 & 0.00 & 0.11 & 0.16 \\
$o_{ds4,c}$ & 0.00 & 0.00 & 0.02 & 0.10 & 0.00 & 0.00 \\
$u_{kimi,c}$ & 0.00 & 0.29 & 0.00 & 0.00 & 0.10 & 0.22 \\
$o_{kimi,c}$ & 0.06 & 0.00 & 0.02 & 0.11 & 0.00 & 0.00 \\
\bottomrule
\end{tabular}\end{center}
\noindent\textbf{Step 6. Distance, cost penalty, total score.} $D_m{=}\sqrt{\sum_c(u_{m,c}^{2}+\lambda\,o_{m,c}^{2})}$ (with $\lambda{=}0.35$) and $J_m{=}D_m{+}\beta\,c_m$ (with $\beta{=}0.63$). The routing-math cost scalar $c_m$ comes from §\ref{sec:pool}; the realised per-call USD cost $a_m^{\$}$ is shown alongside for reference:
\vspace{-2pt}\begin{center}\scriptsize\setlength{\tabcolsep}{4pt}
\begin{tabular}{l r r r r r c}
\toprule
\textbf{model} & $c_m$ & $a_m^{\$}$ & $D_m$ & $\beta c_m$ & $J_m$ & \textbf{sel.} \\
\midrule
\mdlogo{qwen}\,\texttt{qwen} & 0.10 & \$0.001386 & 0.908 & 0.063 & 0.971 &  \\
\mdlogo{ds4}\,\texttt{ds4}   & 0.40 & \$0.002895 & 0.562 & 0.252 & 0.814 &  \\
\mdlogo{kimi}\,\texttt{kimi} & 0.60 & \$0.030703 & 0.380 & 0.378 & 0.758 & $\checkmark$ \\
\bottomrule
\end{tabular}\end{center}
\vspace{4pt}\noindent\textbf{Step 7. Decision.} $m^{\star}=\arg\min_m J_m = \mdlogo{kimi}\,\texttt{kimi2.6}$ ($J_{\text{qwen}}{=}0.971$, $J_{\text{ds4}}{=}0.814$, $J_{\text{kimi}}{=}0.758$). The dominant capability is \texttt{creative\_synthesis} at $p{=}0.53$, and $\tau_q{=}0.80$ inflates the per-capability requirement; the qwen capacity on that dimension falls short, so its under-capacity penalty exceeds the cost gap to kimi.

\begin{table}[H]
\centering
\caption{Two additional Brick decisions across capability dimensions, with the dominant capability probability and complexity-derived difficulty.}
\label{tab:example_queries}
\small
\begin{tabular}{l l r l}
\toprule
\textbf{Dimension} & \textbf{Dominant cap.} & $\tau_q$ & \textbf{Selected} \\
\midrule
\texttt{coding} & \texttt{crea}{=}0.66 & 0.56 & \mdlogo{qwen}\,\texttt{qwen} \\
\texttt{planning\_agentic} & \texttt{pln}{=}0.57 & 0.72 & \mdlogo{kimi}\,\texttt{kimi} \\
\bottomrule
\end{tabular}
\end{table}

\subsection{Routing math}\label{sec:brick-math}
Before the formalism, the routing decision in one sentence: each model is a point in a six-dimensional capability space, each query is a region in the same space, and the router picks the cheapest model whose point lies inside (or close enough to) the query's region. The math below makes ``close enough to'' precise via asymmetric residuals and adds a cost term so that the chosen model is not just geometrically closest but also dollar-minimal.

Let $\mathcal{C}{=}\{c_1,\dots,c_D\}$ denote the capability basis with $D{=}6$ (coding, creative\_synthesis, instruction\_following, math\_reasoning, planning\_agentic, world\_knowledge), and let $\mathcal{M}{=}\{m_1,\dots,m_K\}$ be the model pool of size $K{=}3$ (§\ref{sec:pool}). Each model $m$ has a non-negative cost scalar $a_m$ (the dimensionless USD ratio of §\ref{sec:pool}) and a vector of per-capability success rates $s_{m,c}\in(0,1)$, collected in a skill matrix $S\in(0,1)^{K\times D}$.

\paragraph{Skill matrix calibration.} Each entry $s_{m,c}$ is a probability-weighted empirical success rate estimated offline on the \texttt{results} subset of Dataset~A:
\[
  s_{m,c} \;=\; \mathrm{clip}\!\Bigl(\,\frac{\sum_q p_c(q)\,\mathbb{I}[m\text{ ok on }q]}{\sum_q p_c(q)},\;0.02,\;0.98\Bigr),
\]
where $p_c(q)$ is the ModernBERT capability mass that query $q$ places on dimension $c$ and $\mathbb{I}[m\text{ ok on }q]\in\{0,1\}$ reads the per-model correctness flag from Dataset~A's \texttt{results} subset. No target-model weights are updated; $S$ is locked into the production configuration after calibration.

Two design choices motivate this form. First, the assignment of a query to a capability is \emph{soft}: every query contributes to every capability cell with weight $p_c(q)$, exploiting the full ModernBERT distribution instead of forcing a hard $\arg\max_c p_c(q)$ that would discard the tail of $p(x)$ and waste signal on multi-capability queries. Second, the clip range $[0.02,\,0.98]$ is a fixed calibration convention that keeps $\mathrm{logit}(s_{m,c})$ finite: without it, a cell where the model never (or always) succeeds would send $\mathrm{logit}(s_{m,c})$ to $\pm\infty$ and blow up the capacity term $v_{m,c}$ downstream. On Dataset~A the empirical $s_{m,c}$ values stay comfortably inside the clip band, so the clip acts as a numerical safety net rather than a regularizer that biases the estimate. When the pool is extended to models with very low support on rare capabilities (sparse cells where the clip would start biting), a Bayesian-smoothed variant of this estimator is the natural next step; we discuss it explicitly in §\ref{sec:future-estimator}.

\paragraph{Per-query scoring.} For an incoming query, the ModernBERT classifier (§\ref{sec:brick-modernbert}) produces a capability distribution $p(x){=}(p_{c_1},\dots,p_{c_D})$ and the complexity head produces a blended difficulty $\tau_q\in(0,1)$. We map difficulty to an \emph{effective difficulty logit}
\[
  z_q \;=\; b \;+\; \mu\,\mathrm{logit}(\tau_q),
\]
where $b$ is an additive bias and $\mu$ a multiplicative slope (both set by the user-preference knob, §\ref{sec:brick-knob}). The logit transform is not cosmetic: a raw $\tau_q$ in probability space saturates near the boundaries and treats $0.50{\to}0.55$ as equivalent to $0.94{\to}0.99$, even though the second interval separates failing from oracle-level behaviour. Mapping difficulty to log-odds removes that saturation and places it on the \emph{same additive scale} as the per-model skill $\mathrm{logit}(s_{m,c})$ in the capacity term below, so the residual $r_{q,c}{-}v_{m,c}{=}p_c(z_q{-}\mathrm{logit}(s_{m,c}))$ reads as a per-capability \emph{log-odds gap}.

The affine form $b{+}\mu\,\mathrm{logit}(\tau_q)$ is the minimum-sufficient parametrization on top of that: $\mu$ is a sensitivity ($\mu{\to}0$ removes the dependence on the difficulty classifier while capacity and cost terms stay active), $b$ is a baseline floor, and these are the two scalar handles the preference knob $r$ modulates in §\ref{sec:brick-knob}.

We name $z_q$ explicitly rather than inlining it: $\tau_q$ is the classifier's measurement, $z_q$ is the operator-calibrated quantity consumed by the math. The knob acts on $(\mu, b)$, never on $\tau_q$ itself. Naming $z_q$ separately also avoids attaching per-capability lifting parameters and inflating the calibration surface from two scalars to $2D$.

The \emph{per-capability requirement} is
\[
  r_{q,c} \;=\; p_c\,z_q,
\]
read as ``how hard the query is on capability $c$, weighted by how much that capability matters for this query''. The \emph{per-model per-capability capacity} is
\[
  v_{m,c} \;=\; p_c\,\mathrm{logit}(s_{m,c}),
\]
the analogous quantity for the model. Both quantities carry the same factor $p_c$ on purpose: it makes the gap $r_{q,c}{-}v_{m,c}{=}p_c(z_q{-}\mathrm{logit}(s_{m,c}))$ proportional to how much capability $c$ actually matters for this query, so a dimension with $p_c{=}0$ contributes nothing to the distance no matter how badly any model fails on it. This is the structural difference between Brick and a domain-router: instead of picking one capability and discarding the rest, we form a continuous weighted comparison over all six dimensions and let the math discount the irrelevant ones automatically.

We then decompose the gap into two asymmetric residuals:
\[
  u_{m,c} \;=\; \max\!\bigl(0,\; r_{q,c}-v_{m,c}\bigr),
\]
\[
  o_{m,c} \;=\; \max\!\bigl(0,\; v_{m,c}-r_{q,c}\bigr),
\]
the \emph{under-capacity} $u$ (model is too weak on $c$) and the \emph{over-capacity} $o$ (model is overkill on $c$). We split the residual into two non-negative halves instead of penalizing the symmetric $(r_{q,c}{-}v_{m,c})^2$ because the two failure modes carry asymmetric consequences. Under-capacity translates directly into a higher chance of a wrong answer (the model lacks the skill the query asks for); over-capacity is mostly wasted money and latency (the model could have been smaller). We want the routing score to weight these two regimes independently, which the symmetric absolute residual cannot express. They combine in a weighted Euclidean capability distance and a cost-penalized score:
\[
  D_m = \sqrt{\textstyle\sum_c\!\left(u_{m,c}^2 + \lambda\,o_{m,c}^2\right)},\;
  J_m = D_m + \beta\,a_m,
\]
where $D_m$ is the \emph{capability distance} (a pure quality term: the log-odds gap between what the query needs and what model $m$ offers, aggregated over the six capabilities), $J_m$ is the \emph{routing objective} (the cost function the router actually minimizes, in the optimization-theory sense of an objective to be driven to its argmin), $\lambda\!\ge\!0$ is the \emph{over-capacity penalty} (smaller $\lambda$ tolerates overkill, recovering the standard quality-only objective in the limit $\lambda{\to}0$), and $\beta\!\ge\!0$ is the \emph{cost coefficient} (larger $\beta$ trades quality for price). The router selects $m^{\star}{=}\arg\min_m J_m$, i.e. the model whose quality gap plus cost penalty is smallest; with $\beta{=}0$ this collapses to $\arg\min_m D_m$ alone (quality-only oracle), and with $\beta{\to}\infty$ it collapses to $\arg\min_m a_m$ (cheapest model regardless of quality). In production, ties within a calibrated band $|J_i{-}J_j|<\varepsilon_\tau$ (default $\varepsilon_\tau{=}0.03$) are broken deterministically by ExpectedSuccess $\sum_c p_c\,s_{m,c}$ (higher is preferred), then by lower cost $a_m$; the $\arg\min$ rule above is the unbanded idealized form. See \texttt{brickrouting/router.go} for the exact implementation.

A few more design choices are worth unpacking.

\paragraph{Why Euclidean aggregation.} The aggregation across capabilities is Euclidean ($L_2$) rather than $L_1$ (sum of absolute gaps) or $L_\infty$ (worst gap). $L_1$ would treat every per-capability gap as interchangeable currency and ignore how the deficit is distributed across dimensions; $L_\infty$ would discard every capability except the single worst one. $L_2$ keeps all six dimensions in play and penalizes large residuals super-linearly, so a single sharp under-capacity gap weighs more than the same total mass split across several small ones, which matches the empirical observation that a model failing badly on the dominant capability of a query is harder to recover than one that is mildly off across the board.

\paragraph{Why $\lambda$ multiplies only $o^2$.} The asymmetric weight $\lambda$ rescales only the over-capacity branch, so the two regimes can be tuned independently: pushing $\lambda{\to}0$ recovers the standard distance-to-floor objective in which overkill is free, while increasing $\lambda$ progressively punishes wasted capability (typically a proxy for cost or latency we cannot price directly).

\paragraph{Why the cost penalty is additive.} The term $\beta\,a_m$ is added linearly to $D_m$ rather than multiplied because the two quantities live in incommensurable units (a log-odds gap and a dimensionless dollar ratio), and a multiplicative form would force them to share scale before any trade-off has even been defined. The additive form treats $\beta$ as the explicit conversion rate from cost units into distance units, so operators can read off how much extra log-odds gap they are willing to accept per dollar of saving.

\paragraph{Why these four scalars are not hyperparameters.} The four scalars $(\mu, b, \beta, \lambda)$ are not free knobs sitting in a config file but functions of the preference knob $r\in[-1,1]$ exposed at routing time (§\ref{sec:brick-knob}).

\begin{figure*}[!t]
\centering
\includegraphics[width=\textwidth]{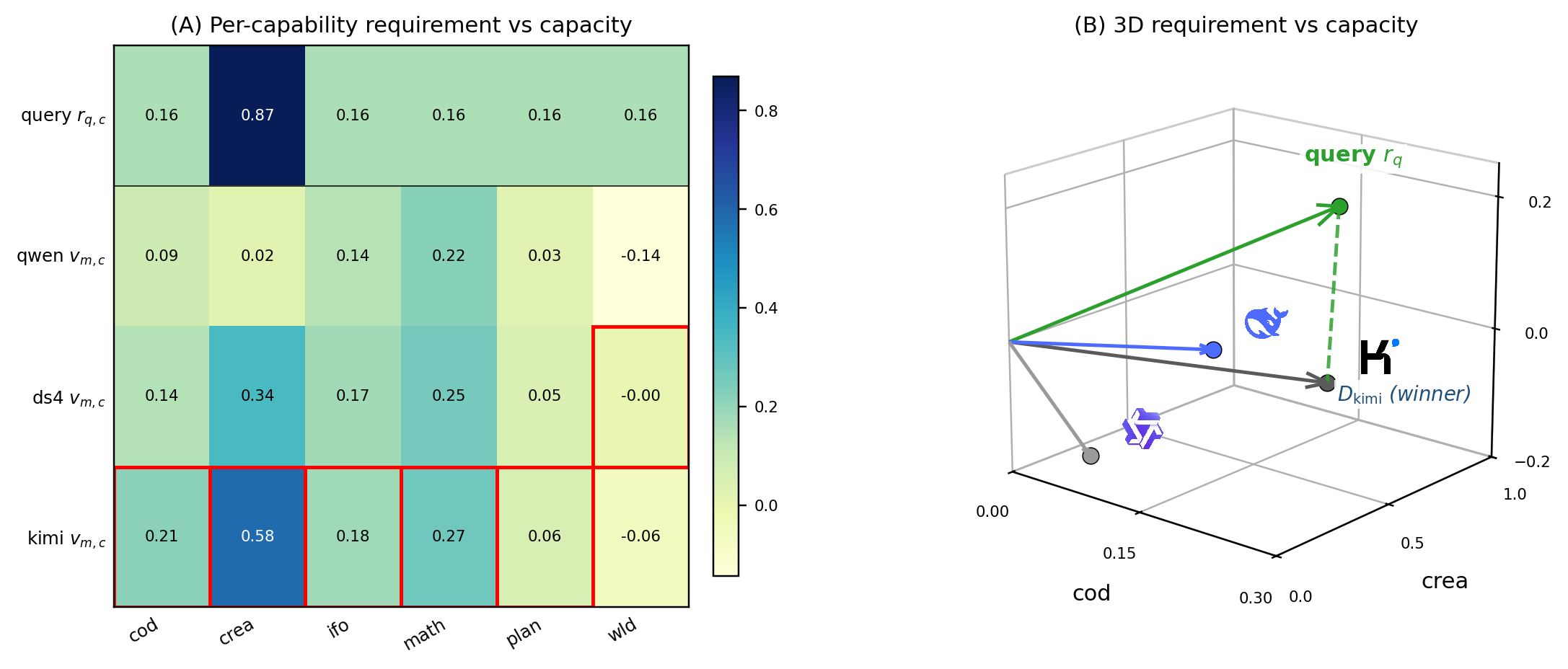}
\caption{Two views of the routing decision for the worked-example query \texttt{q\_03563} of §\ref{sec:brick-example}. \textbf{(A)} Heatmap of the post-projection quantities the router actually compares: the top row is the per-capability requirement $r_{q,c}{=}p_c\,z_q$ (Step~3), the next three rows are the per-model capacities $v_{m,c}{=}p_c\,\mathrm{logit}(s_{m,c})$ (Step~4); the red box in each column marks the model whose capacity is closest to the query requirement on that dimension. \textbf{(B)} 3D projection of the same four vectors onto three of the six axes (\texttt{cod}, \texttt{crea}, \texttt{wld}): the green arrow is the query requirement, the three other arrows are the model capacities (logos at the vector tips), and the dashed arc marks the winning capability distance $D_\text{kimi}{=}0.380$ (Step~6). Both panels use the numerical values reported in Steps 1--6 of §\ref{sec:brick-example}.}
\label{fig:cap_views}
\end{figure*}

Figure~\ref{fig:cap_views} shows two complementary views of the routing decision for the worked example of §\ref{sec:brick-example}.

\subsection{ModernBERT capability classifier}\label{sec:brick-modernbert}
Step (iii) is a fine-tuned ModernBERT classifier~\cite{warner2024modernbert} that maps a query string to a probability vector over six capability dimensions. We fine-tune \texttt{answerdotai/ModernBERT-base} on Brick2 Dataset~B (\texttt{brick2-dataset-b-modernbert-train}), a 50k-query labeled corpus distinct from Brick2 Dataset~A. Training is a W\&B~sweep over learning rate ($10^{-5}$ to $10^{-4}$, log), weight decay ($10^{-6}$ to $10^{-2}$), warmup ratio ($0.03$ to $0.12$), and number of epochs ($2$ to $5$). The winning configuration is selected by the macro-Pearson correlation on a held-out evaluation split, reaching $0.983$.
\begin{itemize}[noitemsep,topsep=2pt,leftmargin=1em]
  \item \href{https://huggingface.co/datasets/regolo/brick2-dataset-b-modernbert-train}{HuggingFace dataset (Brick2 Dataset B, ModernBERT training)}
  \item \href{https://huggingface.co/regolo/brick-complexity-2-eco}{HuggingFace model (Brick capability classifier)}
\end{itemize}

\begin{figure*}[!t]
\centering
\includegraphics[width=0.98\textwidth]{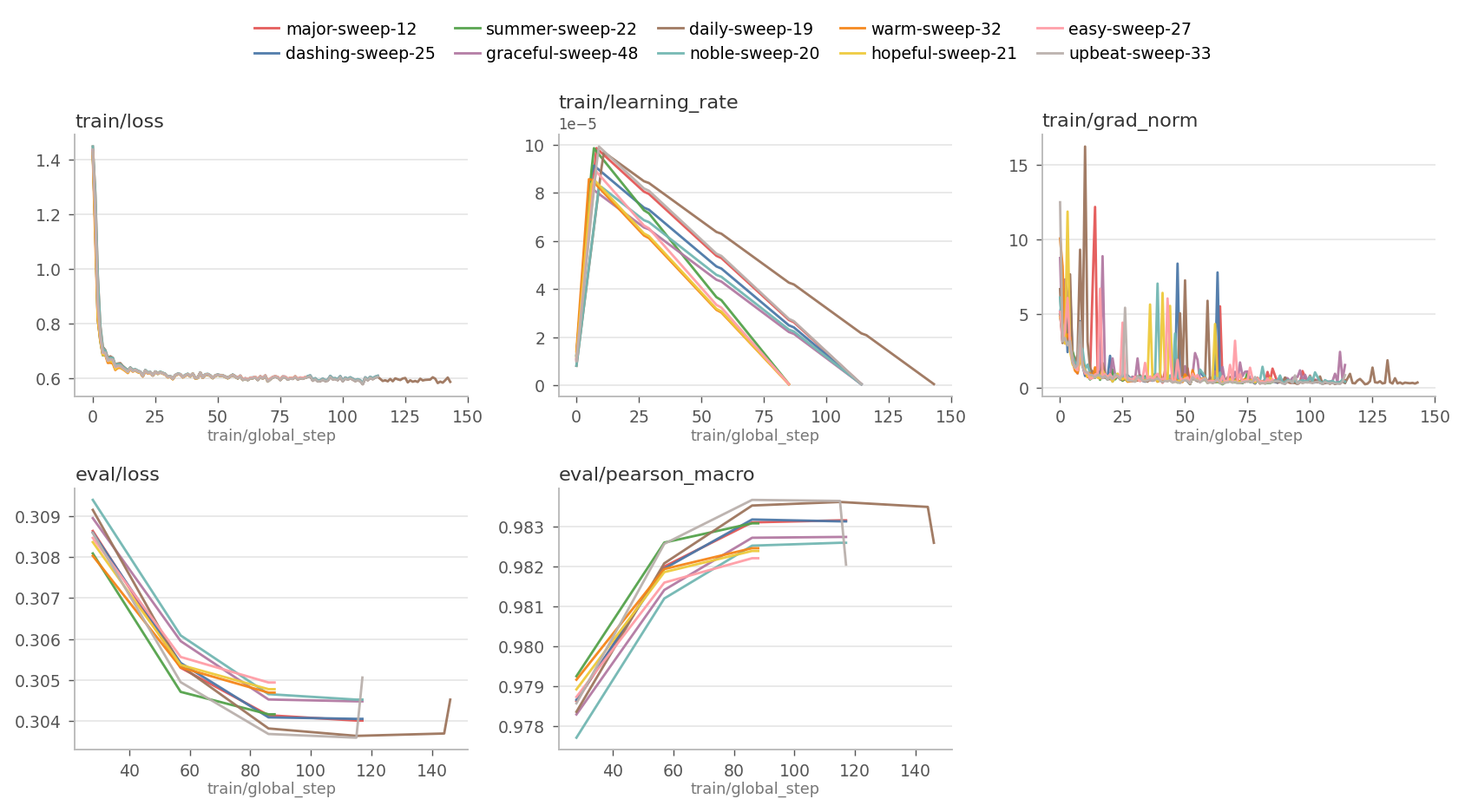}
\caption{Top-10 runs from the ModernBERT W\&B~sweep by Pearson-macro, in a multi-metric dashboard view (training loss, learning rate, gradient norm, validation loss, validation Pearson macro).}
\label{fig:modernbert}
\end{figure*}

\subsection{Complexity classifier and difficulty blending}\label{sec:brick-complexity}
Step (iv) of the pipeline is a separate fine-tuned classifier that scores how hard the query is for the pool. We use a 3-class head (\texttt{easy}/\texttt{medium}/\texttt{hard}) built on \texttt{Qwen3.5-0.8B} with LoRA adapters ($r{=}\alpha{=}32$, dropout $0.1$), fine-tuned for $3$ epochs at learning rate $10^{-4}$, batch size $16$, maximum sequence length $768$. Training uses an asymmetric over-penalty ($\lambda_{\text{over}}{=}0.7$, biasing the classifier to under- rather than over-rate difficulty) and label smoothing $0.08$; on the held-out test set ($n{=}1994$) it reaches $72.8\%$ accuracy and $0.425$ macro-F1. The classifier returns a label $\ell\in\{\text{easy},\text{medium},\text{hard}\}$ and a confidence $c\in[0,1]$.

Three anchor values bind labels to a continuous difficulty axis,
\[
  \tau_{\text{easy}}{=}0.55,\quad \tau_{\text{medium}}{=}0.72,\quad \tau_{\text{hard}}{=}0.88,
\]
and the blended difficulty $\tau_q$ consumed by the router (§\ref{sec:brick-math}) is a confidence-weighted convex combination of the predicted-label anchor and the medium anchor:
\[
  \tau_q \;=\; c\cdot\tau_{\ell} \;+\; (1-c)\cdot\tau_{\text{medium}}.
\]
When the classifier is uncertain ($c{\to}0$) the difficulty collapses to the safe medium tier ($\tau_q{\to}\tau_{\text{medium}}$); when it is confident ($c{\to}1$) it matches the predicted anchor. This blending replaces a brittle hard decision with a smooth contraction toward the centre, absorbing the residual miscalibration of the small classifier without escalating to a larger model just to break a tie. As a concrete example, the query of §\ref{sec:brick-example} is labelled \texttt{hard} with $c{=}0.51$, yielding $\tau_q = 0.51\cdot 0.88 + 0.49\cdot 0.72 = 0.802$.
\begin{itemize}[noitemsep,topsep=2pt,leftmargin=1em]
  \item \href{https://huggingface.co/regolo/brick-complexity-2-eco}{HuggingFace: regolo/brick-complexity-2-eco}
\end{itemize}

\paragraph{Domain-agnosticism.} The complexity head is intentionally domain-agnostic: it consumes the raw query string and does not condition on the capability distribution $p(x)$ produced by the ModernBERT head upstream. A domain-conditioned variant that takes $p(x)$ (or the dominant capability $\arg\max_c p_c$) as auxiliary input is a natural next iteration and could materially improve difficulty calibration on capability-specific failure modes; we leave this extension to future work and discuss it in §\ref{sec:discussion}.

\subsection{User preference knob}\label{sec:brick-knob}
Brick exposes a continuous preference $r\in[-1,1]$ with named profiles $\mathrm{min}{\equiv}{-1}$, $\mathrm{neutral}{\equiv}0$, $\mathrm{max}{\equiv}{+1}$. The knob smoothly modulates the four routing scalars $(\mu(r),\,b(r),\,\beta(r),\,\lambda(r))$ that enter the scoring of §\ref{sec:brick-math}. Concretely, define the positive and negative branches
\[
  p^{+}(r) \;=\; \max(r,0)^{\alpha}, \qquad p^{-}(r) \;=\; \max(-r,0)^{\alpha},
\]
where $\alpha{>}0$ is a sharpness exponent (larger $\alpha$ flattens the response near $r{=}0$ and concentrates the action at the extremes). The four scalars follow a calibrated asymmetric power law:
\[
\begin{aligned}
  \mu(r)     &\;=\; \mu_0\,\exp\!\bigl(p^{+}\ln A_{\mu}^{+} \;+\; p^{-}\ln A_{\mu}^{-}\bigr), \\
  b(r)       &\;=\; b_0 \;+\; p^{+}A_{b}^{+} \;+\; p^{-}A_{b}^{-}, \\
  \beta(r)   &\;=\; \beta_0\,\exp\!\bigl(-p^{+}\ln A_{\beta}^{+} \;+\; p^{-}\ln A_{\beta}^{-}\bigr), \\
  \lambda(r) &\;=\; \lambda_0\,\exp\!\bigl(-p^{+}\ln A_{\lambda}^{+} \;+\; p^{-}\ln A_{\lambda}^{-}\bigr).
\end{aligned}
\]
The four base values $(\mu_0, b_0, \beta_0, \lambda_0)$ are the neutral-knob ($r{=}0$) defaults; the eight $A_{\cdot}^{\pm}$ are asymmetric multipliers that let calibration push harder on one side of the knob than the other. The construction is not arbitrary; below we trace where it comes from, then justify each design choice in turn.

\paragraph{Where these formulas come from.} Three constraints shape the family: continuity in $r$, identity at $r{=}0$ (neutral reproduces production defaults), and pre-calibrated targets at $r{=}\pm 1$. These constraints are met by many continuous families; we pick a one-sided power law $|r|^\alpha$ as the simplest interpretable parametrization that allows independent asymmetric calibration of the two extremes, hence $|r|^\alpha$ as the building block. Since the two extremes are not mirror images (e.g.\ $\lambda$ must shrink by ${\approx}900\times$ at $r{=}{+1}$ but inflate only ${\approx}12\times$ at $r{=}{-1}$), we split into one-sided basis functions $p^{+}(r){=}\max(r,0)^\alpha$ and $p^{-}(r){=}\max(-r,0)^\alpha$: both vanish at $0$, exactly one is unity at $\pm 1$, never overlap. The combiner is then dictated by domain: $\mu, \beta, \lambda$ must stay positive, so $x_0\cdot\exp(\cdot)$; $b$ lives in logit space, so $b_0+(\cdot)$. We wrap the calibration constants in $\ln$ inside the exponent purely for readability: at $r{=}\pm 1$ the exponent collapses to $\pm\ln A$, so $A^\pm$ is literally the multiplicative factor at each extreme (e.g.\ $A_\lambda^{+}{=}896$ reads off the table as ``$\lambda$ shrinks $896\times$ at max-quality''). The sign in front of each $p^{\pm}\ln A^{\pm}$ is fixed by routing semantics: at $r{=}{+1}$ we want $\beta, \lambda$ to shrink (minus sign) and $\mu$ to grow (plus sign); $b$ stays straight additive.

\paragraph{Why two one-sided branches $p^{+}, p^{-}$ instead of a signed $|r|^\alpha$.} A signed term would still vanish at $r{=}0$ but would force a single constant to govern both extremes. We wanted independent calibration per side because the production trade-offs are not symmetric: for example, the over-capacity penalty $\lambda$ has to deflate by a factor of about $50$ at $r{=}{+1}$ (so overkill stops mattering when the operator picks max-quality) and inflate by a factor of about $1{,}000$ at $r{=}{-1}$ (more than enough to push the router toward cheap models), while the cost coefficient $\beta$ requires a $6{,}559{\times}$ deflation at $r{=}{+1}$ and only a $9{\times}$ inflation at $r{=}{-1}$. That ratio of ratios cannot be expressed with a single multiplier; the split into $p^{+}, p^{-}$ with independent $A^{+}, A^{-}$ is the cheapest parametrization that buys per-side freedom while keeping continuity and intrinsic neutrality at $r{=}0$.

\paragraph{Combiner form follows from each scalar's domain.} $\mu, \beta, \lambda$ are positive magnitudes (scaling a logit, an additive cost, a sum of squares respectively), so we wrap them as $x_0\cdot\exp(\pm p^{\pm}\ln A)$ which gives $x_0\cdot A^{\mp 1}$ at $r{=}{\pm 1}$ and lets calibration constants be read off the table directly. The bias $b$ instead enters as $z_q = b + \mu\,\mathrm{logit}(\tau_q)$ and has the dimension of a logit (real-valued, sign-bearing), so the natural form is additive: $A_b^{\pm}$ then expresses how many units of log-odds the baseline moves at each extreme (e.g.\ $A_b^{-}{=}{-}1.35$ subtracts about $1.4$ log-odds at $r{=}{-1}$, the magnitude bounded by the calibration constraint $|A_b^{-}|{\le}5.0$ to ensure cost-monotonic behaviour across the knob range).

\paragraph{Why the signs on $p^{+}, p^{-}$ are inverted for $\beta, \lambda$.} By convention $r{=}{+1}$ is the max-quality profile: in that regime we want the cost coefficient $\beta$ and the over-capacity penalty $\lambda$ to be \emph{small} so the router stops worrying about price and about overkill. The negative sign in front of $p^{+}\ln A_{\beta}^{+}$ produces $\beta_0 / A_{\beta}^{+}$ at $r{=}{+1}$ (with $A_{\beta}^{+}{=}6559$ that is $\beta \approx 0.23 / 6559 \approx 3.5{\times}10^{-5}$: cost becomes effectively free, the router chases quality); the positive sign in front of $p^{-}\ln A_{\beta}^{-}$ produces $\beta_0 \cdot A_{\beta}^{-}$ at $r{=}{-1}$ (with $A_{\beta}^{-}{=}8.8$ that is $\beta \approx 0.23 \cdot 8.8 \approx 2.0$: cost dominates, the router collapses to the cheapest model). The same construction holds for $\lambda$. For $\mu$ the sign convention is the opposite: $r{=}{+1}$ should \emph{inflate} the difficulty signal so larger models qualify, hence the $+p^{+}\ln A_{\mu}^{+}$ in the exponent. The $b$ branch is simpler: $r{=}{+1}$ pushes $b$ upward by $A_b^{+}$ (queries look harder) and $r{=}{-1}$ pulls $b$ downward by $A_b^{-}$ (queries look easier).

The locked production values are listed in Table~\ref{tab:brick-knob-defaults}. The five base scalars $(\mu_0, b_0, \beta_0, \lambda_0, \tau_\text{base})$ were jointly tuned via Bayesian optimization on a held-out development split of Dataset~A (search space and winning values in Appendix~A.1, Table~\ref{tab:app_searchspace}). The exponent $\alpha$ and the eight power-law multipliers $A_{\bullet}^{\pm}$ were then obtained by a warm-start sweep that scored each candidate configuration on the full $5{,}504$-query five-fold out-of-fold curve under a multi-signal objective: peak accuracy at $r{=}{+}1$, monotonic cost across the five preference profiles, and per-dimension dispatch signals (e.g.\ neutral-profile \texttt{ds4} share on \texttt{world\_knowledge}). The $|A_b^{-}|{\le}5.0$ constraint mentioned above was added to this second-stage search to prevent a bias-collapse pathology at mid-band negative $r$ that produced a non-monotonic cost curve.

At inference, the knob is passed per-request via the HTTP header \texttt{X-Brick-Routing-Preference} (a float in $[-1,1]$) or the convenience header \texttt{X-Brick-Routing-Profile} with named values \texttt{eco}/\texttt{balanced}/\texttt{pro} mapping to $r{=}{-1},0,{+1}$. The thirteen power-law constants are loaded once from the service configuration at startup and never recomputed at routing time.

\begin{table*}[!t]
\centering
\caption{Locked production math configuration for the Brick preference knob (symbols above, values below).}
\label{tab:brick-knob-defaults}
\small
\setlength{\tabcolsep}{5pt}
\begin{tabular}{*{13}{c}}
\toprule
$\mu_0$ & $b_0$ & $\beta_0$ & $\lambda_0$ & $\alpha$ &
$A_{\mu}^{+}$ & $A_b^{+}$ & $A_{\beta}^{+}$ & $A_{\lambda}^{+}$ &
$A_{\mu}^{-}$ & $A_b^{-}$ & $A_{\beta}^{-}$ & $A_{\lambda}^{-}$ \\
\midrule
$0.345$ & $0.82$ & $0.231$ & $0.045$ & $2.92$ &
$13.0$ & $5.29$ & $6559$ & $49.5$ &
$0.081$ & $-1.35$ & $8.8$ & $1002$ \\
\bottomrule
\end{tabular}
\end{table*}

\section{Brick Results on Dataset A}\label{sec:results}

\paragraph{Mechanism.} The headroom that lets an MoM router exceed the best single model is structural, not numerical. \texttt{kimi2.6} is not the per-query oracle: it does not solve $1{,}375$ of the $5{,}504$ queries, of which $453$ are correctly answered by at least one cheaper model. The two-model oracle restricted to \texttt{kimi}$+$\texttt{ds4} reaches $81.96\%$ ($+6.94$\,pp over \texttt{kimi} alone), and the full \emph{three-model} oracle on $\{$\texttt{qwen},\texttt{ds4},\texttt{kimi}$\}$ reaches $83.25\%$ ($+8.23$\,pp over \texttt{kimi}); the latter is the true ceiling an MoM router can in principle extract over the strongest single baseline on this pool. Brick's $+1.96$\,pp lift is the fraction of that headroom actually captured by the cost-penalized geometric dispatch rule. Two structural effects drive it. First, \emph{complementary error sets}: \texttt{deepseek-v4-flash} and \texttt{qwen3.5-9b} correctly answer queries that \texttt{kimi2.6} either declines or fails, with the \texttt{world\_knowledge} refusal pattern being the cleanest case ($51.7\%$ refusal rate for \texttt{kimi2.6} vs sub-$10\%$ for the other two). This refusal behaviour is qualitatively different from a hallucinated wrong answer: \texttt{kimi2.6} opts for honest conservatism (``I don't know'' or empty response) where the other two would attempt and sometimes succeed (see §\ref{sec:discussion}, ``Honest refusal as an exploitable model property''). The router therefore is not just patching wrong answers, it is also patching strategic refusals, and Brick's skill vectors capture this asymmetry per capability dimension. Second, \emph{dimension-weighted dispatch}: the capability vector $p(x)$ lets the router act on a model's \emph{specific} weak capability rather than its global average, so a query loading mostly on a dimension where the strongest model has a softness is rerouted automatically. A third effect specific to the mid-band profiles is \emph{deepseek-v4-flash sweet-spot exploitation}: on the \texttt{world\_knowledge} capability \texttt{deepseek-v4-flash} reaches $49\%$ accuracy, exceeding \texttt{kimi2.6}'s $34\%$ (and far above \texttt{qwen}'s $18\%$); the locked Brick parameters dispatch $50$--$53\%$ of low/neutral-profile queries to \texttt{ds4}, absorbing the bulk of the under-routing failures (queries dispatched to a cheap model that cannot solve them, where \texttt{ds4} can) that an aggressive cheap-only baseline would incur, at $\approx 10{\times}$ lower cost per call than \texttt{kimi2.6}. The residual gap between Brick ($76.98\%$) and the three-model oracle ($83.25\%$) is the portion of the headroom that the current router does not yet extract; closing it is the optimization direction discussed in §\ref{sec:discussion}.

Table~\ref{tab:brick_oof} reports Brick's full-dataset out-of-fold curve under stratified $5$-fold skill-vector calibration. For each fold, skill vectors are calibrated on the four held-in folds and evaluated on the held-out fold; knob hyperparameters are fixed from prior development. This protocol uses the entire dataset while avoiding per-example leakage in the skill estimates. It should be read as an in-domain calibrated evaluation, not as a zero-shot result.

\begin{table}[H]
\centering
\caption{Brick out-of-fold results on Dataset~A under locked knob hyperparameters. Selected-answer accuracy (Sel.\,acc.) is the fraction of queries where the dispatched model solved the task. Dispatch percentages are rounded to the nearest $1\%$ (rows may not sum exactly to $100\%$).}
\label{tab:brick_oof}
\small
\begin{tabular}{l r r r r}
\toprule
\textbf{Profile} & $r$ & \textbf{Sel.\,acc.} & \textbf{Avg.\,cost (USD)} & \textbf{Dist.\,(\%)} \\
\midrule
min     & $-1.0$           & $63.17\%$          & $\$0.001386$ & $100/0/0$ \\
low     & $-0.5$           & $71.62\%$          & $\$0.003557$ & $45/51/5$ \\
neutral & $\phantom{-}0.0$ & $74.11\%$          & $\$0.006513$ & $29/56/15$ \\
high    & $+0.5$           & $76.24\%$          & $\$0.014905$ & $21/35/44$ \\
max     & $+1.0$           & $\mathbf{76.98\%}$ & $\$0.022083$ & $0/31/69$ \\
\bottomrule
\end{tabular}
\end{table}

The per-fold skill vectors used to produce Table~\ref{tab:brick_oof} are computed with the estimator of §\ref{sec:brick-math}, evaluated independently on each held-in fold so that no per-query correctness signal from the held-out fold enters the skill matrix.

The max-quality profile reaches $76.98\%$, exceeding always-\texttt{kimi} ($75.02\%$) by $+1.96$ pp while spending $\$0.022083$ vs $\$0.030703$ per call ($\approx \mathbf{28\%}$ cheaper, i.e.\ $1.39{\times}$). The neutral profile sits at $74.11\%$ ($+10.94$ pp over always-\texttt{qwen}) at cost $\$0.006513$, $\mathbf{4.71{\times}}$ cheaper than always-\texttt{kimi} for only $0.91$\,pp accuracy loss; it also exceeds Cascade Routing ($73.40\%$ at $\$0.006113$) by $+0.71$\,pp on accuracy at near-parity cost. It dispatches $56\%$ of queries to \texttt{deepseek-v4-flash} and only $15\%$ to \texttt{kimi}, exploiting the empirical fact that \texttt{ds4} outperforms \texttt{kimi} on the \texttt{world\_knowledge} capability dimension (see Mechanism above). The low profile reaches $71.62\%$ at $\$0.003557$, exceeding FrugalGPT's $69.42\%$ accuracy by $+2.20$\,pp at $\approx 1.27{\times}$ FrugalGPT's accepted-cost and at $8.63{\times}$ lower cost than always-\texttt{kimi}: this operating point dominates the cascade family in the low-cost regime on response accuracy, achieved by dispatching $51\%$ to \texttt{deepseek-v4-flash} and only $5\%$ to \texttt{kimi} on the queries that \texttt{qwen} cannot solve. The min profile reaches always-\texttt{qwen}'s $63.17\%$ at $\$0.001386$, $\mathbf{22.15{\times}}$ cheaper than always-\texttt{kimi}. The gap from the max profile to the $83.25\%$ three-model oracle is the residual routing opportunity not captured by the current six-dimensional capability features and the deterministic decision rule.

\section{Latency Analysis}\label{sec:latency}
We report latency along two axes: the raw \emph{router overhead} (decision time, LLM call excluded) and the \emph{perceived end-to-end latency} (router decision plus LLM generation time).

\paragraph{Measurement protocol.} All systems were evaluated on the same deterministic ordering of the $5{,}504$ queries (grouped by capability dimension, no per-system shuffle), so any API-side load variation is correlated across rows rather than across systems. Each query enters the latency statistics once: values are \emph{single-pass} measurements, not averaged over repeated runs. We did not perform a within-system variance analysis; the reported figures are single-pass measurements under public-API conditions. The full $5{,}504$-query test set is administered to every system without subsetting, filtering, or cherry-picking; no outlier exclusion is applied, and we retain every query that returns a non-null, strictly positive latency value. The only discarded entries are those with missing or non-positive latency records (network-level transport failures), which would distort the distribution if folded in and which we do not selectively re-introduce to favour any system.


\begin{table}[H]
\centering
\caption{Router decision overhead (LLM call excluded) on Dataset A ($N{=}5{,}504$). All values in milliseconds.}
\label{tab:router_overhead}
\small
\setlength{\tabcolsep}{4pt}
\begin{tabular}{l r r r r}
\toprule
\textbf{Router} & \textbf{p50} & \textbf{p95} & \textbf{p99} & \textbf{mean} \\
\midrule
Brick (ours)            & 2998 & 3211 & 3311 & 1883 \\
Cascade Routing         & 15.1 & 19.7 & 48.6 & 16.0 \\
FrugalGPT               & 103.3 & 157.7 & 556.9 & 90.2 \\
RouteLLM binary         & 97.5 & 138.3 & 394.6 & 84.0 \\
RouteLLM tournament     & 197.8 & 306.1 & 1318 & 175.2 \\
\bottomrule
\end{tabular}
\end{table}

\begin{table}[H]
\centering
\caption{Perceived end-to-end latency (router decision plus LLM generation) on Dataset A. The MoM (Brick max profile) row joins Brick's router latency with the LLM latency of the selected model per query. Single-model baselines show LLM generation latency only.}
\label{tab:latency_perceived}
\small
\begin{tabular}{l r r r r}
\toprule
\textbf{System} & \textbf{p50} & \textbf{p95} & \textbf{p99} & \textbf{mean} \\
\midrule
always-qwen   & 21441 & 232232 & 266774 & 52677 \\
always-ds4    & 10169 & 135296 & 468067 & 41768 \\
always-kimi   & 51246 & 390276 & 738106 & 106926 \\
\textbf{Brick (MoM)}      & \textbf{22802} & \textbf{278621} & \textbf{820288} & \textbf{78178} \\
\bottomrule
\end{tabular}
\end{table}

The router-overhead numbers in Table~\ref{tab:router_overhead} reflect the Dataset~A benchmark harness, where Brick makes two auxiliary classifier calls (capability and complexity) over a remote endpoint, while Cascade Routing performs a local hash lookup. In a production deployment of Brick observed over $6{,}376$ live requests, the same decision step has a median of $494$ ms (p95 $2{,}528$ ms), reflecting warm caches and co-located classifiers. The benchmark overhead is therefore a conservative upper bound on practical decision latency.

\begin{figure}[!t]
\centering
\includegraphics[width=\columnwidth]{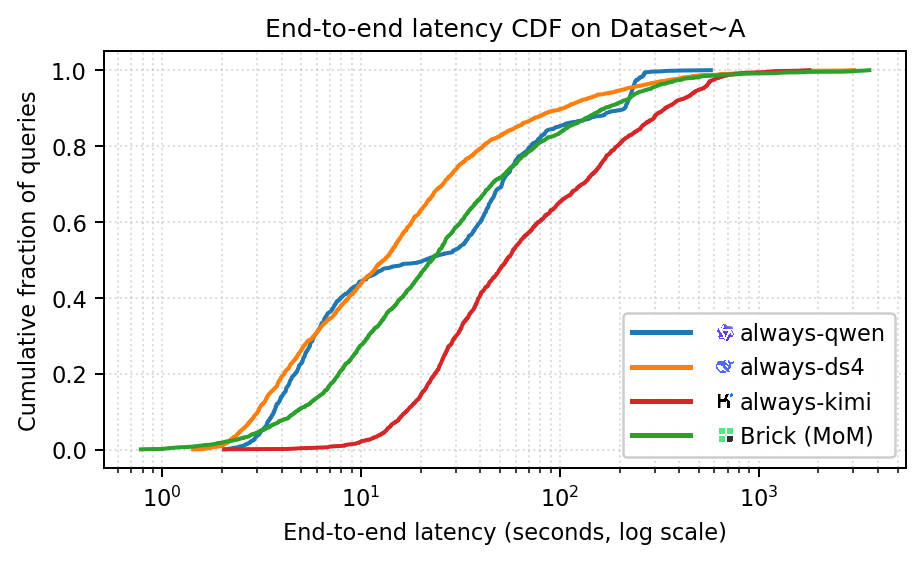}
\caption{End-to-end latency CDF on Dataset~A. Brick (MoM) at the max profile sits between always-\texttt{ds4} and always-\texttt{kimi}, while achieving higher response accuracy than either.}
\label{fig:latency_cdf}
\end{figure}

For perceived latency (Table~\ref{tab:latency_perceived}), Brick at the max-quality profile reaches the same response-accuracy regime as always-\texttt{kimi} at less than half its median end-to-end latency ($22.8$ s vs $51.2$ s). The reason is that even at the max profile Brick routes roughly $31\%$ of queries to \texttt{deepseek-v4-flash}, which is materially faster than \texttt{kimi2.6} per generation. Figure~\ref{fig:latency_cdf} shows the latency CDFs: Brick's distribution sits to the left of always-\texttt{kimi} at every quantile, and rejoins always-\texttt{ds4} at the median.

\section{Discussion and Limitations}\label{sec:discussion}
The reported results support three claims. (i) MoM dispatches queries across whole models and is a viable design point distinct from MoE. (ii) Within a three-model pool, a calibrated capability-aware router (Brick) exceeds the best single-model baseline on response accuracy at lower cost and lower perceived latency. (iii) The headroom above the best router and below the oracle ($83.25\%{-}76.98\%{=}6.27$ pp) reflects information the router does not yet extract from the query and the model skill matrix; this is the direction where future MoM routers will compete.

\paragraph{Honest refusal as an exploitable model property.} On the \texttt{world\_knowledge} dimension of Dataset~A, \texttt{kimi2.6} explicitly refuses to answer (``I don't know'' or empty response) on $51.7\%$ of queries, versus $6.6\%$ for \texttt{qwen3.5-9b} and $8.1\%$ for \texttt{deepseek-v4-flash}; on creative-synthesis the three models refuse at comparable rates. Refusal is graded as incorrect by the judge, but it is qualitatively different from hallucination: \texttt{kimi2.6} opts for honest conservatism rather than a confident wrong answer. Once skill vectors capture per-capability refusal rates, the router naturally avoids dispatching factual queries to a model that would prefer to refuse.

The current evaluation is in-domain calibrated: the skill matrix is fit on held-in folds of Dataset~A, and the knob hyperparameters were selected on Dataset~A. Strict external generalization should be measured on a fully out-of-distribution benchmark before claims are extended to arbitrary deployment workloads. The reported latency numbers assume a public-API deployment for the LLM call and a remote classifier endpoint for the router; numbers will move under self-hosted inference.

Extending to mixed open-weight and closed-weight pools is the principal open direction: the router treats commercial APIs as additional points in the cost-capability plane, with skill-vector calibration and non-stationary pricing as the remaining engineering problems.

\paragraph{Why pure routing wins for agents.}\label{sec:agentic-discussion} Pure single-step routing is not just a stylistic preference over cascade routing: it is the only design that scales to agentic workloads without compounding cost and latency on every step. On our pool the measured per-decision overhead of a cascade router is $1.47\times$ (FrugalGPT) above the accepted-model cost; the strictly sequential variant of Cascade Routing pays an analogous $1.43\times$ overhead under the same probe-and-escalate schedule (§\ref{sec:baselines-cascade}). Because every step of an agent loop independently re-runs the cascade gate from the cheapest tier, this overhead is paid $N$ times on an $N$-step trajectory: by $N{=}10$ the cascade bill is $43\%$ to $47\%$ higher than that of a pure router operating at the same accepted-model cost (Table~\ref{tab:agent_step_blow_up}), and the gap is linear in $N$. Latency is hit in the same way, because cascade stages are sequential by construction (a downstream gate must observe the rejected model's output before it can fire), so every rejected stage adds a full round-trip to the user's critical path. For agents that issue tool calls or planner-executor steps in the tens, this turns into seconds of perceived idle time per request, on top of the inflated bill. A pure router (Brick included) pays the routing decision once per step and forwards a single LLM call; the overhead is bounded by the classifier cost (sub-second in production, see §\ref{sec:latency}) and does not scale with cascade depth. We consider this the strongest practical argument for pure MoM routing in production agent stacks.

\subsection{Open directions for the estimator}\label{sec:future-estimator}
Three directions naturally extend the current calibration. They are concrete enough to define a research agenda, and none of them requires touching the routing math: each is a drop-in upgrade to a single component of the estimator stack.

\paragraph{Bayesian skill estimator for sparse pools.} The probability-weighted estimator of §\ref{sec:brick-math} relies on a clip $[0.02,\,0.98]$ as a numerical safety net. On Dataset~A's three-model pool this is harmless because every $(m,c)$ cell has support in the thousands, so the empirical mean lands comfortably inside the band. When the pool is extended (say, adding a fourth or fifth model with a narrow capability profile, or a closed-weight model with sparse public-benchmark coverage), some cells will land near the boundaries with thin support and the clip will start biasing the estimate. A Beta-Binomial smoothed variant is the principled replacement:
\[
  s_{m,c} \;=\; \frac{K_{m,c} + k\,\mu_c}{N_{m,c} + k}, \qquad k = 8,
\]
where $K_{m,c}$ counts correct answers under hard arg-max assignment, $N_{m,c}$ is the cell support, $\mu_c$ is the cross-model accuracy prior on capability $c$, and $k$ is the pseudo-count. Sparse cells then fall back to the typical pool performance $\mu_c$ rather than to a hard floor, and the estimator converges to the empirical mean on cells with abundant support. The router config already exposes a \texttt{prior\_strength} field (default $8.0$) as the slot for this estimator; wiring the smoother through the production calibrator and re-running the five-fold sweep is left to future work.

\paragraph{Domain-conditioned complexity estimation.} A second structural open direction sits inside the router itself. As noted in §\ref{sec:brick-complexity}, the current complexity head is intentionally domain-agnostic: it consumes the raw query and does not condition on the capability distribution $p(x)$. A domain-conditioned variant that takes $p(x)$ as auxiliary input could materially sharpen difficulty calibration on capability-specific failure modes (e.g., math-vs-world-knowledge difficulty are not on the same scale even at equal token length); we expect this to be the next iteration of the difficulty estimator and the most likely route to closing a non-trivial fraction of the residual oracle gap.

\paragraph{Vision: query entropy as a thermodynamic quantity.} The capability distribution $p(x){\in}\Delta^{D-1}$ produced by the ModernBERT head is, formally, a probability distribution over the six-dimensional capability simplex. This invites a thermodynamic reading of the routing problem. Queries with near-uniform $p(x)$ are ``hot'', spreading mass across many capabilities and asking the router to balance gaps along several axes at once; queries with peaked $p(x)$ are ``cold'', concentrating all mass on one capability and reducing the decision to an essentially one-dimensional comparison. The preference knob $r$ acts as a temperature in the same vocabulary: at $r{=}{+}1$ the cost penalty collapses and the router behaves as if it had ``free energy'' to spend on the capability the query loads on, while at $r{=}{-}1$ the cost term dominates and routing collapses onto the cheapest model regardless of capability shape. A formal study of query entropy as a single-scalar query feature, of the partition-function-like aggregation of the per-model distances $D_m$ across the pool, and of a free-energy decomposition of the routing objective $J_m$ into capability and cost contributions, is a direction we plan to pursue. Beyond the conceptual appeal, the practical hope is a principled, knob-free tie-breaker between models of comparable $J_m$: one that selects the model minimizing a temperature-weighted combination of expected accuracy and expected cost, in place of the current hand-tuned $\varepsilon_\tau$ band. We see this as the most speculative of the three directions and the one most likely to reshape the routing math itself rather than a single component on top of it.

\section{Conclusion}\label{sec:conclusion}
Whole-model routing across heterogeneous LLM pools, framed as \emph{Mixture of Models}, is a deployment pattern worth treating as a first-class object. Brick instantiates it as a calibrated capability-aware router that exceeds the best single-model baseline and external routing baselines on Dataset~A at competitive latency, and serves as a practical bridge for operators mixing open-weight and commercial closed-weight models behind a single product.

Model-level routing also extends along the \emph{modality} axis: adding a vision- or audio-capable model with a modality-detection signal lifts a text-only deployment into an effectively multimodal service without retraining the text backbone.

\appendix
\section{Hyperparameter Selection Protocol}\label{app:hpo}

\paragraph{Two classes of hyperparameters.} The router exposes two distinct classes of internal constants: \emph{(i) calibrated parameters}, jointly tuned on a held-out development split of Dataset~A via Bayesian optimization, and \emph{(ii) design-time constants}, fixed at implementation time and not part of the optimization budget. We document both classes below for full transparency; a sensitivity analysis is reported only for class (i), since class (ii) was not jointly searched. The motivation for this two-class split is operational: class (i) parameters control the cost-quality trade-off the operator cares about at deploy time, while class (ii) parameters express structural choices (the difficulty scale, the skill-matrix clip, the cost normalization) that we expect to stay stable across deployments.

\subsection*{A.1 Class (i): Bayesian-tuned parameters}
Six numerical knobs were jointly searched via Weights\&Biases Bayesian optimization (\texttt{method: bayes}, hyperband early-termination), with the dev/test split fixed at \texttt{dev\_fraction = 0.70} (test held-out for §\ref{sec:results}). Across four sweep tables we logged $\sim 25{,}000$ configurations.

\begin{table}[H]
\centering
\caption{Class-(i) Bayesian search space and winning configuration. Distributions: $\log$ = log-uniform, $\mathrm{U}$ = uniform.}
\label{tab:app_searchspace}
\small\setlength{\tabcolsep}{4pt}
\begin{tabular}{l c c c c}
\toprule
\textbf{Parameter} & Sym. & Range & Distr. & Win. \\
\midrule
Fallback difficulty       & $\tau_\text{base}$ & $[0.30,\,0.97]$ & $\mathrm{U}$  & $0.80$ \\
Difficulty lift mean      & $\mu$              & $[0.05,\,2.0]$  & $\log$        & $0.35$ \\
Difficulty lift bias      & $b$                & $[-1.5,\,1.5]$  & $\mathrm{U}$  & $0.82$ \\
Cost penalty              & $\beta$            & $[5{\times}10^{-3},\,2.0]$ & $\log$ & $0.23$ \\
Over-capacity penalty     & $\lambda$          & $[2{\times}10^{-3},\,2.0]$ & $\log$ & $0.05$ \\
User preference knob      & $r$                & $[-1,\,1]$      & $\mathrm{U}$  & $+1.00$ \\
\bottomrule
\end{tabular}
\end{table}

\subsection*{A.2 Selection criterion and metric reconciliation}
The Bayesian sweep optimizes \texttt{holdout\_accuracy}, defined as the rate at which the router selects the \emph{cheapest correct} model on the $30\%$ held-out partition: a query answered correctly by all three pool models has a \texttt{qwen} ground-truth label, so routing to \texttt{kimi2.6} counts as wrong even when \texttt{kimi2.6} also produces a correct response. This is a deliberately strict metric, designed to expose cost-aware routing efficiency rather than end-user quality. The headline number $76.98\%$ reported in §\ref{sec:results} is instead \texttt{response\_accuracy}, the fraction of queries on which the \emph{selected} model produces a correct response, evaluated on the full $5{,}504$ queries at the max-quality preference setting $r{=}{+}1$. The two metrics measure different aspects of the same router and are not in tension: \texttt{holdout\_accuracy} guides the calibration to a cost-balanced operating point, while \texttt{response\_accuracy} is the deploy-time figure operators see end-to-end.

\subsection*{A.3 Class (ii): Design-time constants (not jointly tuned)}
The following constants are fixed at implementation time and were \emph{not} part of the Bayesian sweep:
\begin{itemize}\setlength{\itemsep}{3pt}
  \item \textbf{Difficulty anchors} $\tau_\text{easy}{=}0.55$, $\tau_\text{medium}{=}0.72$, $\tau_\text{hard}{=}0.88$ (§\ref{sec:brick-complexity}): three uniformly-spaced anchors on the $[0,1]$ difficulty axis, with $\tau_\text{medium}{=}0.72$ chosen to approximately match the observed \texttt{deepseek-v4-flash} dev-set accuracy ($\approx 0.74$). These are deployment defaults specified once in the Brick reference math and are not jointly searched with the class-(i) parameters.
  \item \textbf{Skill-matrix probability-weighted estimator} (§\ref{sec:brick-math}): the production calibration uses a soft, probability-weighted average bounded by the clip range below; a Bayesian-smoothed variant with pseudo-count $k{=}8$ is the natural extension for pools with sparse $(m,c)$ cells and is discussed as future work in §\ref{sec:future-estimator}.
  \item \textbf{Power-law shape constants} of the preference-knob mapping (the exponent $\alpha{=}2.92$ and the eight multipliers $A_{\bullet}^{\pm}$): tuned by a warm-start sweep on top of the class-(i) optimum that scored each candidate on the full $5{,}504$-query five-fold OOF curve under a multi-signal objective (peak accuracy at $r{=}{+}1$, cost-monotonicity across the five profiles, per-dimension dispatch signals), subject to the constraint $|A_b^{-}|\!\le\!5.0$ to enforce a monotonic cost curve.
  \item \textbf{Routing-math cost scalar} $c_\text{qwen}{=}0.10$, $c_\text{ds4}{=}0.40$, $c_\text{kimi}{=}0.60$: dimensionless normalization matched to public per-1M-output-token price ratios at the time of evaluation, on a scale comparable to the routing distance $D_m$ (§\ref{sec:pool}); locked at calibration time, not searched. The real per-call dollar cost $a_m^{\$}$ used in result tables is a derived quantity computed from observed token volumes, not a routing-math input.
  \item \textbf{Skill-matrix clip range} $[0.02,\,0.98]$ on $s_{m,c}$: a fixed convention to keep logit transforms finite; not tuned.
\end{itemize}
Jointly searching class (ii) alongside class (i) is a natural extension and is left to future work; we expect the largest available gain to come from joint calibration of the anchor triple with the difficulty classifier.

\subsection*{A.4 Sensitivity analysis on class (i)}
Marginal sensitivity of \texttt{holdout\_accuracy} around the winning configuration, aggregated over the $\sim 25{,}000$ sweep rows: for each parameter we bin runs into $\pm 20\%$ and $\pm 50\%$ neighbourhoods around the winning value (other parameters left free), and report the mean shift in \texttt{holdout\_accuracy}. The worst-case column shows the largest mean-of-bin degradation observed across the full searched range.

\begin{table}[H]
\centering
\caption{Sensitivity of \texttt{holdout\_accuracy} to perturbations of the class-(i) parameters around the winning configuration. Values in percentage points (pp). The $r$ knob is a deployment-time control and not noise.}
\label{tab:app_sensitivity}
\scriptsize\setlength{\tabcolsep}{3.5pt}
\begin{tabular}{l c c c c c}
\toprule
\textbf{Param} & Win & $\Delta$acc $\pm 20\%$ & $\Delta$acc $\pm 50\%$ & Worst & V. \\
\midrule
$\tau_\text{base}$     & $0.45$ & $-2.24$ & $-2.49$ & $-5.20$@$0.72$ & fragile \\
$\mu$                  & $0.25$ & $+1.13$ & $+1.01$ & $-0.99$        & moderate \\
$b$                    & $0.30$ & $+1.25$ & $+1.25$ & $-1.47$        & moderate \\
$\beta$                & $0.02$ & mixed   & $-0.48$ & $-3.62$@$0.40$ & fragile \\
$\lambda$              & $0.20$ & $+0.76$ & $-0.07$ & $-0.95$        & robust \\
$r$                    & $-0.50$ & deploy knob & n/a & n/a & control \\
\bottomrule
\end{tabular}
\end{table}

Two parameters ($\tau_\text{base}$ and $\beta$) are sensitive to large perturbations and require care at deployment; the remaining four sit within $\pm 1.5$\,pp under $\pm 20\%$ perturbation. The two sensitive parameters are precisely the ones with the most direct effect on the difficulty-vs-cost trade-off (the base difficulty fallback and the cost penalty), which is consistent with their role: they are the parameters by which the operator dials the trade-off, so a flat response would have been the surprise.

\paragraph{Reproducibility checklist.} The protocol above addresses items related to training/test details and hyperparameter selection commonly required by ML reproducibility checklists (NeurIPS, ICLR): the search space, optimization method, selection metric, and sensitivity analysis are reported explicitly, and class-(ii) constants are listed with an honest acknowledgment that they were not part of the joint search. We do not claim the configuration is a global optimum; we report it as the Pareto-optimal point that emerged from the searched budget, under the metric chosen for cost-balanced operation.

\section*{Credits}
We thank Seeweb for providing the GPU infrastructure (a $4{\times}$NVIDIA L40S cluster) used to host the open-weight model pool and the ModernBERT capability and complexity classifier endpoints throughout calibration and evaluation. We thank Regolo.ai for providing the model-serving stack and the production deployment of Brick from which the latency figures of §\ref{sec:latency} were measured. Brick is currently available as an experimental meta-model on the Regolo.ai platform under the public identifier \texttt{brick-v1-beta}, where readers can reproduce the routing behaviour described in this paper on their own queries.

\begingroup
\footnotesize\selectfont

\endgroup

\end{document}